\documentclass{article} 
\usepackage{scidiagbench_preprint,times}
\usepackage{float}

\usepackage{amsmath,amsfonts,bm}

\def\eqref#1{equation~\ref{#1}}

\def\1{\bm{1}}

\DeclareMathAlphabet{\mathsfit}{\encodingdefault}{\sfdefault}{m}{sl}
\SetMathAlphabet{\mathsfit}{bold}{\encodingdefault}{\sfdefault}{bx}{n}

\usepackage{hyperref}
\usepackage{url}
\usepackage{graphicx}
\usepackage{booktabs}
\usepackage{amssymb}
\usepackage{multirow}
\usepackage{tabularx}
\usepackage[most]{tcolorbox}

\definecolor{resourcelinkblue}{HTML}{163B80}

\definecolor{promptpurple}{HTML}{7C3AED}
\definecolor{promptpurplelight}{HTML}{F5F0FF}
\definecolor{promptpurpleheader}{HTML}{E9D5FF}
\newtcolorbox{definitionbox}{
  enhanced,
  breakable,
  colback=promptpurplelight,
  colframe=promptpurple,
  colbacktitle=promptpurpleheader,
  coltitle=black,
  fonttitle=\bfseries,
  title=Definition,
  boxrule=0.8pt,
  arc=2pt,
  left=7pt,right=7pt,top=5pt,bottom=5pt,
  before skip=4pt,after skip=7pt
}
\newtcolorbox{criterionbox}[1]{
  enhanced,
  breakable,
  colback=white,
  colframe=promptpurple,
  colbacktitle=promptpurpleheader,
  coltitle=black,
  fonttitle=\bfseries,
  title=#1,
  boxrule=0.8pt,
  arc=2pt,
  left=7pt,right=7pt,top=5pt,bottom=5pt,
  before skip=4pt,after skip=7pt
}

\title{SciDiagBench: A Type-Specific Benchmark for Human-Aligned Evaluation of Scientific Diagrams}

\author{
\textbf{Hao Wang}\textsuperscript{1,2,5},
\textbf{Zixuan Li}\textsuperscript{1,2,3,*},
\textbf{Long Bai}\textsuperscript{1,2,3},
\textbf{Xiaolong Jin}\textsuperscript{1,2,4,*} \\
\textbf{Jiafeng Guo}\textsuperscript{1,2,4},
\textbf{Tat-Seng Chua}\textsuperscript{3},
\textbf{Tian Tian}\textsuperscript{5} \\[0.6em]
{\small\textsuperscript{1}Institute of Computing Technology, Chinese Academy of Sciences} \\
{\small\textsuperscript{2}State Key Laboratory of AI Safety} \\
{\small\textsuperscript{3}National University of Singapore} \\
{\small\textsuperscript{4}School of Computer Science, University of Chinese Academy of Sciences} \\
{\small\textsuperscript{5}School of Advanced Interdisciplinary Sciences, University of Chinese Academy of Sciences} \\[0.35em]
{\small\hypersetup{hidelinks}\texttt{%
\href{mailto:wanghao24@mails.ucas.ac.cn}{wanghao24@mails.ucas.ac.cn}\quad
\href{mailto:lizixuan@ict.ac.cn}{lizixuan@ict.ac.cn}\quad
\href{mailto:jinxiaolong@ict.ac.cn}{jinxiaolong@ict.ac.cn}}} \\[0.6em]
{\small\hypersetup{hidelinks}%
\setlength{\tabcolsep}{0pt}%
\begin{tabular}{@{}c@{\hspace{0.4em}}l@{\hspace{0.4em}}l@{}}
  \raisebox{-0.2em}{\includegraphics[height=1.2em]{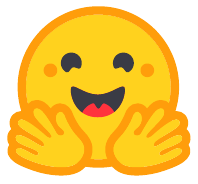}} &
  \textbf{Leaderboard:} &
  \href{https://huggingface.co/spaces/golaxy/SciDiagBench-Leaderboard}{\textcolor{resourcelinkblue}{\nolinkurl{huggingface.co/spaces/golaxy/SciDiagBench-Leaderboard}}} \\[0.25em]
  \raisebox{-0.2em}{\includegraphics[height=1.2em]{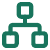}} &
  \textbf{Explorer:} &
  \href{https://xiaotang404.github.io/SciDiagBench-Explorer/}{\textcolor{resourcelinkblue}{\nolinkurl{xiaotang404.github.io/SciDiagBench-Explorer/}}}
\end{tabular}}
}
\date{}

\hypersetup{
  pdftitle={SciDiagBench: A Type-Specific Benchmark for Human-Aligned Evaluation of Scientific Diagrams},
  pdfauthor={Hao Wang, Zixuan Li, Long Bai, Xiaolong Jin, Jiafeng Guo, Tat-Seng Chua, Tian Tian}
}

\makeatletter
\newcommand{\firstpagecorrespondence}{%
  \begingroup
  \renewcommand{\thefootnote}{\fnsymbol{footnote}}%
  \def\Hy@footnote@currentHref{scidiagbench.correspondence}%
  \footnotetext[1]{Corresponding authors: Zixuan Li and Xiaolong Jin.}%
  \endgroup
}
\makeatother

\newcommand{\benchname}{\textsc{SciDiagBench}}

\begin{document}

\maketitle
\firstpagecorrespondence
\begin{abstract}
Recent agents have demonstrated strong capabilities in generating images from textual prompts, motivating their utilization for scientific diagram generation. Scientific diagrams naturally span diverse types, each characterized by distinct visual structures and content organization. Consequently, during peer review and other scientific evaluation, human experts naturally consider type-specific criteria when assessing their quality.
However, existing VLM-based evaluation benchmarks fail to reflect the above human evaluation practices for two main reasons: 1) benchmark datasets typically lack a systematic categorization of different diagram types and thus adopt the same generic criteria for all diagrams, and 2) their evaluation criteria are often designed heuristically, without explicit alignment to human evaluation preferences.
Motivated by these limitations, we introduce \benchname{}, which consists of a benchmark dataset that systematically categorizes scientific diagrams into 11 types and a human preference-guided evaluation method. Specifically, \benchname{} contains 165 scientific diagrams across 10 broad domains, with types including graphical abstracts, technical roadmaps, and illustrative cases. Based on this dataset, \benchname{} develops an evaluation method that leverages a preference-guided calibration process to incorporate human preferences into the generation of type-specific criteria. Ultimately, \benchname{} enables more reliable automated evaluation that better aligns with human evaluation across diverse scientific diagrams.
\end{abstract}

\section{Introduction}
\label{sec:intro}
Scientific diagrams, as an important complement to scientific texts, play a crucial role in communicating complex technical concepts through visual representations \citep{zhu2026autofigure}. However, crafting high-quality diagrams remains labor-intensive, requiring both domain expertise and design skills. As AI-assisted research workflows increasingly extend from hypothesis generation and experimentation to scientific reporting and visualization \citep{xu2026ai4science}, automating diagram creation has become a natural next step. Building on advances in multimodal generative models (e.g., Gemini 3 Pro Image \citep{google2025gemini3proimage} and GPT-Image-2 \citep{openai2026gptimage2}), recent studies have developed scientific diagram generation agents, such as PaperBanana \citep{zhu2026paperbanana} and AutoFigure \citep{zhu2026autofigure}. These agents enable the direct processing of long scientific manuscripts to extract essential information and synthesize target diagrams by coordinating multiple stages of diagram creation.

As scientific diagram generation agents become increasingly capable, a critical question is how to reliably evaluate the quality of generated diagrams in practical scientific settings, such as peer review. Compared with general images, scientific diagrams are particularly challenging to evaluate, as they span diverse types designed for different communicative purposes. Consequently, human experts naturally apply different evaluation criteria when assessing different types of scientific diagrams, making type-specific evaluation essential. For example, while technical roadmaps place greater emphasis on structural criteria to ensure protocol completeness and logical correctness, illustrative cases prioritize content criteria to faithfully represent data instances and execution traces. 

On the dataset side, rather than systematically organizing diagrams according to their communicative purposes and visual structures, existing datasets often define their categories based on disciplinary domains or specific generation tasks. For example, benchmarks centered on artificial intelligence primarily cover diagrams such as methodological frameworks and flowcharts \citep{mondal2024scidoc2diagrammer,zhu2026paperbanana}, while multidisciplinary benchmarks typically organize their samples by scientific domains rather than by diagram types \citep{chang2025sridbench,luo2025mmmg}. As a result, they provide limited support for evaluation that accounts for the diverse quality requirements of different scientific diagrams and reflects human evaluation practices.

On the evaluation side, manual assessment of scientific diagrams is often labor-intensive and time-consuming, motivating recent efforts toward automated evaluation. Current evaluation methods generally fall into two main paradigms. The first paradigm derives criteria from individual reference diagrams. For instance, some studies evaluate generated diagrams by matching extracted structural information \citep{luo2025mmmg,chang2026fepbench} or by applying tailored checklists \citep{li2026easier,zhao2026kvbench,li2026bizgeneval,sun2026t2ireasonbench,pang2025paper2poster} specific to each reference. While offering fine-grained assessments, these diagram-specific methods are difficult to generalize across instances and scale efficiently. The second paradigm, in contrast, applies a shared set of criteria across all diagrams, typically constructed from predefined dimensions and heuristic rules \citep{zhu2026paperbanana,zhu2026autofigure,mondal2024scidoc2diagrammer,pang2025paper2poster}. Although this approach facilitates large-scale evaluation, these overly generic criteria lack explicit alignment with the evaluation principles and preferences of human experts.

To address these limitations, we introduce \benchname{}, a comprehensive benchmark designed for automated evaluation of scientific diagram generation. Specifically, we collect scientific diagrams from diverse scholarly sources and construct a systematic taxonomy based on their visual structures, content organization, and communicative purposes, yielding 165 diagrams across 11 types and 10 domains. 
Building on this dataset, we develop an evaluation framework that generates type-specific criteria across multiple quality dimensions, refines the criteria with scoring instructions derived from human preferences, and adopts reference-based scoring for stable and comparable evaluation.
Utilizing \benchname{}, we systematically evaluate four representative agents (Vanilla, PaperBanana, AutoFigure, and Paper2Figure). Extensive analysis reveals a highly uneven capability profile: while current agents excel at organizing abstract conceptual relationships, they exhibit systematic deficits in generating complex topologies and precise concrete visual elements, highlighting critical bottlenecks in existing scientific diagram generation agents.

Our primary contributions are summarized as follows:

\begin{itemize}
    \item We construct the first type-specific benchmark dataset for scientific diagram generation, with 11 diagram types and 10 scientific domains.

    \item We develop a human preference-guided evaluation framework that generates type-specific criteria and further refines them with scoring instructions derived from human preferences.

    \item We conduct extensive experiments on representative agents using \benchname{}, revealing substantial variations across agents, diagram types, and judge models. Human evaluation further demonstrates that the proposed automated evaluation framework achieves better alignment with human judgments.
\end{itemize}

\section{Related Work}
\label{sec:related_work}

\paragraph{Scientific Diagram Generation}
\label{related_work:sci_diag_gen}
Scientific diagram generation methods broadly fall into two categories: code-based generation and image-based generation. Code-based approaches formulate diagram creation as a text-to-code task, typically using languages such as TikZ~\citep{belouadi2025tikzero} or Python~\citep{ge2025autopresent,zheng2025pptagent} to produce structured and editable representations. Recent systems further incorporate iterative reasoning and refinement, as in SciDoc2Diagrammer-MAF~\citep{mondal2024scidoc2diagrammer} and SciFig~\citep{huang2026scifig}. However, code-based approaches are often limited in visual expressiveness when handling complex scientific content~\citep{zhu2026paperbanana}. In contrast, image-based approaches directly synthesize diagrams with multimodal generative models (e.g., Gemini 3 Pro Image~\citep{google2025gemini3proimage} and GPT-Image-2~\citep{openai2026gptimage2}). Building on increasingly capable image generators, recent agents such as AutoFigure~\citep{zhu2026autofigure}, PaperBanana~\citep{zhu2026paperbanana}, and Paper2Figure (the scientific-figure-generation module of Paper2Any)~\citep{liu2025paper2any} introduce explicit content planning and intermediate representations to improve generation quality. Despite these rapid advancements, the systematic assessment of agent outputs remains challenging.

\paragraph{Evaluation of Generated Diagrams}
Conventional metrics such as CLIPScore~\citep{hessel2021clipscore}, Fr\'{e}chet Inception Distance (FID)~\citep{heusel2017gans}, and CLIP Maximum Mean Discrepancy (CMMD)~\citep{jayasumana2024rethinking} measure image--text alignment or distributional similarity and are insensitive to errors in diagram entities and their logical relations. Meanwhile, using agents as judges has become a new paradigm for evaluating open-ended generation, evolving from static scoring toward dynamic evaluation~\citep{shi2026judgeagent}. Building on this paradigm, recent studies adopt more structured evaluation protocols for image generation, typically supported by Vision-Language Models (VLMs). One line of work develops prompt-based benchmarks that assess images generated from carefully constructed prompts. T2I-CoreBench~\citep{li2026easier}, KVBench~\citep{zhao2026kvbench}, and BizGenEval~\citep{li2026bizgeneval} associate each prompt with atomic checklists. T2I-ReasonBench assesses reasoning and image quality~\citep{sun2026t2ireasonbench}. To evaluate semantics more precisely, FEPBench employs atomic decomposition~\citep{chang2026fepbench}, while MMMG uses knowledge-graph edit distance~\citep{luo2025mmmg}. Another line of work develops paper-based benchmarks to evaluate the end-to-end generation from scientific manuscripts. DiagramEval matches nodes and paths in graph representations of Scalable Vector Graphics (SVG) diagrams~\citep{liang2025diagrameval}; SridBench~\citep{chang2025sridbench}, FigureBench~\citep{zhu2026autofigure}, and PaperBananaBench~\citep{zhu2026paperbanana} use source contexts and reference diagrams for shared multidimensional scoring; SciFig adds paper-specific questions to common criteria~\citep{huang2026scifig}.

\section{The SciDiagBench Benchmark}
\label{sec:the_scidiagbench_benchmark}
\begin{figure}[t]
    \centering
    \includegraphics[width=\linewidth]{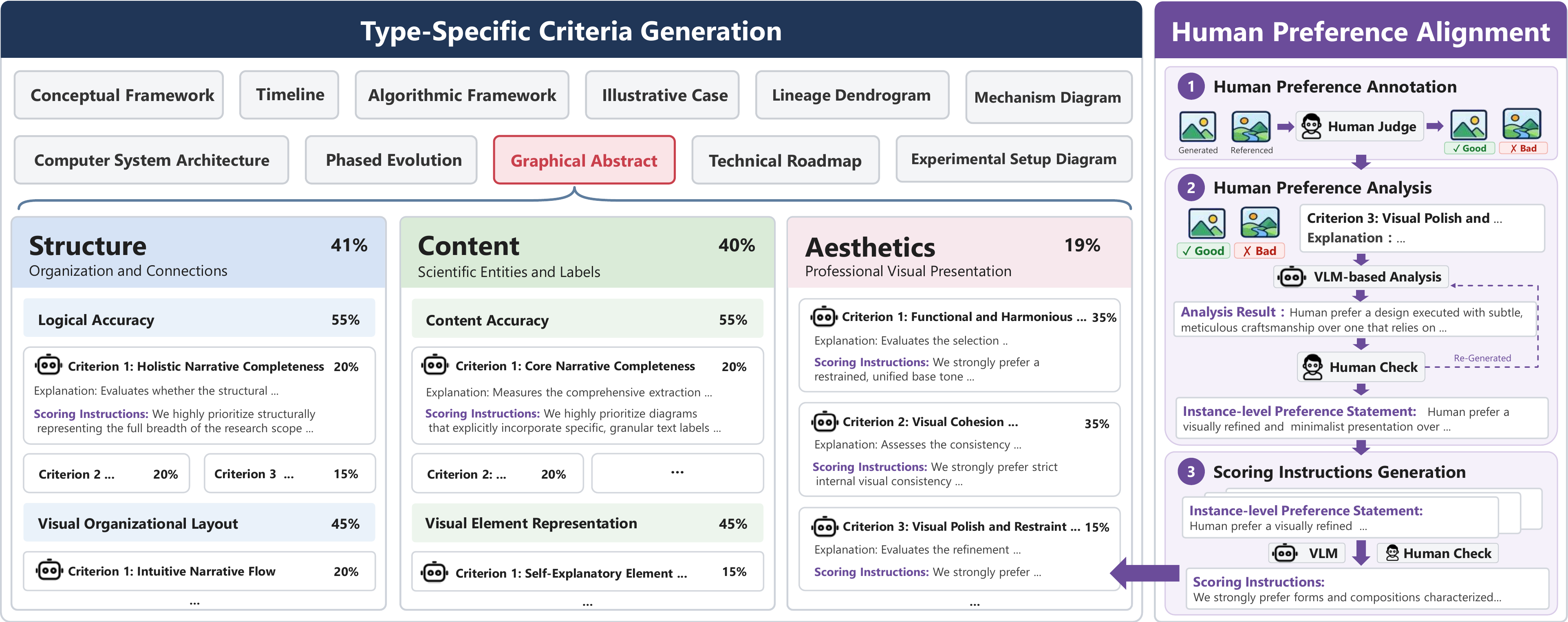}
    \caption{Overview of the dynamic hierarchical evaluation framework, combining type-specific criteria generation with human preference alignment.}
    \label{fig:dynamic-hierarchical-evaluation}
\end{figure}

In general, this section is organized into three main parts: formal task definition, dataset construction pipeline, and design of a customized evaluation framework (an overview of which is illustrated in Figure~\ref{fig:dynamic-hierarchical-evaluation}).

\subsection{Task Definition}
Following PaperBanana~\citep{zhu2026paperbanana}, we formulate the task as the end-to-end generation of a scientific diagram by an agent. Given a scientific paper $P$, a target diagram caption $C$, and a target aspect ratio $R$, an agent $\mathcal{A}$ directly generates the final diagram $\hat{D}$:
\begin{equation}
    \hat{D} = \mathcal{A}(P,C,R).
    \label{eq:task-formulation}
\end{equation}

The task does not prescribe any intermediate representation or generation procedure; instead, it evaluates the end-to-end capability of $\mathcal{A}$ to extract relevant information from $P$ that falls within the scope specified by $C$ and generate a clear, coherent, and publication-ready diagram under the specified aspect ratio constraint $R$.

\subsection{Dataset Construction}

\paragraph{Diagram Collection.}
To provide a foundation for defining the task scope and developing the diagram taxonomy, it is essential to ensure the diversity of visual representations. 
We initially collect 364 papers from diverse sources such as interdisciplinary and field-specific scholarly journals (e.g., \emph{The Innovation} and \emph{Reviews of Modern Physics}), major international conferences (e.g., ICIS), and the open-access arXiv preprint repository.
A total of 598 figures are sampled from these papers.
From this initial set, we observe that certain categories are not well suited for direct generation by agents. Specifically, we exclude two broad classes of figures based on their visual representations. Figures in the first class rely on rigid quantitative mappings or outputs produced by specialized tools, including \emph{plots}, \emph{tables}, \emph{data-driven maps}, \emph{diagrams illustrating mathematical and physical principles}, \emph{chemical structure and reaction diagrams}, and \emph{mechanical CAD diagrams}. Figures in the second class require direct sensor-based acquisition, primarily including \emph{photos}~\citep{dhote2023figuresurvey} and composite figures containing real-world images. Collectively, these excluded categories are inherently better suited for generation via deterministic visualization tools, specialized software, or physical sensors. Furthermore, to mitigate potential training-data contamination, we manually inspect the candidate diagrams and filter out those whose preliminary generation output reproduces the reference diagram with anomalously high similarity. Further details on the excluded categories and the contamination filtering process are provided in Appendix~\ref{sec:dataset-construction-details}.
After applying filtering, we retain a total of 291 candidate diagrams for subsequent stages.

\paragraph{Taxonomy Construction.}
Existing taxonomies of scientific figures, such as DocFigure~\citep{jobin2019docfigure} and ACL-FIG-PILOT~\citep{karishma2023aclfig}, are primarily designed for downstream tasks including figure retrieval and classification. Consequently, they typically categorize figures based on broad graphic modalities (e.g., photographs and tables) or quantitative charting formats (e.g., bar charts and line graphs)~\citep{dhote2023figuresurvey}. To comprehensively evaluate scientific diagram generation, we construct a fine-grained taxonomy based on the diverse visual structures, content organization, and communicative purposes of target diagrams. By analyzing the remaining 291 heterogeneous scientific diagrams, we identify recurring combinations of these characteristics and categorize them into 11 types. Specifically, some diagrams are organized along temporal or evolutionary axes: \emph{Timeline} (TL), \emph{Lineage Dendrogram} (LD), and \emph{Phased Evolution} (PE) distinguish chronological sequences of milestones, branching lineage relationships, and transitions among discrete system states, respectively. Other types emphasize overall structural frameworks: \emph{Algorithmic Framework} (AF), \emph{Computer System Architecture} (CSA), and \emph{Conceptual Framework} (CF) illustrate computational modules and their execution dependencies, layered hardware--software components and interfaces, and abstract constructs and their relations, respectively. The remaining types are defined by specific communicative objectives: \emph{Technical Roadmap} (TR) for staged methodological procedures and decision paths, \emph{Graphical Abstract} (GA) for paper-level summaries, \emph{Experimental Setup Diagram} (ESD) for hardware and instrumentation setups, \emph{Illustrative Case} (IC) for explanatory examples, and \emph{Mechanism Diagram} (MD) for causal interactions and state transformations. Detailed definitions and characteristics of each type are provided in Appendix~\ref{sec:appendix-diagram-types}.

\paragraph{Dataset Partitioning.}
To facilitate human preference alignment and support comprehensive evaluation and analyses, we partition the scientific diagrams into a development set and a test set. 
Given the pronounced category imbalance among the remaining 291 scientific diagrams, we first construct a balanced subset by sampling 15 diagrams for each type.
Then, these diagrams are randomly partitioned into a development set (5 per type) and a test set (10 per type), with strict controls to ensure no overlap in source papers. 
Each finalized instance is represented as a quadruple $(P,C,R,D)$, consisting of a scientific paper, a caption, an aspect ratio, and a reference diagram. For the scientific paper $P$, we parse each source PDF using MinerU~\citep{niu2025mineru25}, retaining the main body text while removing author metadata, references, and other irrelevant materials. The target aspect ratio $R$ is derived by mapping the width-to-height ratio of the reference diagram $D$ to the closest standard aspect ratio (e.g., 3:2).

\subsection{Evaluation Framework}
\label{sec:evaluation-framework}

\paragraph{Type-Specific Criteria Generation.}
Unconstrained criteria generation may produce criteria that target irrelevant aspects rather than essential qualities~\citep{du2026deepresearchbench}. We guide this process using three complementary top-level dimensions. \emph{Structure} concerns the spatial organization of a diagram; \emph{Content} concerns the domain-specific information conveyed by its constituent elements; and \emph{Aesthetics} concerns graphic design quality. However, the top-level dimensions alone do not distinguish perceptual execution from semantic validity. A diagram may exhibit a clear layout or visually recognizable elements while containing incorrect dependencies or factual statements. Since visual presentation is salient while logical or factual errors are less immediately perceptible, a VLM judge may conflate the two, leading to overestimated overall scores. We therefore decompose Structure into \emph{Visual Organizational Layout} (VOL) and \emph{Logical Accuracy} (LA), and Content into \emph{Visual Element Representation} (VER) and \emph{Content Accuracy} (CA), while retaining Aesthetics as a single dimension (detailed in Appendix~\ref{sec:appendix-evaluation-dimensions}). Conditioned on these dimensions and the target diagram type $\tau$, we leverage Gemini 3.1 Pro to dynamically generate tailored criteria $\mathcal{C}_{d,\tau}$ for dimension $d$ under diagram type $\tau$, alongside two-level normalized weights comprising intra-dimension weights $w_c$ and top-level dimension weights $W_{d,\tau}$. Notably, criteria for Structure and Content are generated directly under their respective sub-dimensions, but their intra-dimension weights $w_c$ are normalized within the corresponding top-level dimension. Dimension-level weights $W_{d,\tau}$ are averaged over 5 independent trials to improve stability. Each generated criterion is manually reviewed for quality and relevance.

\paragraph{Human Preference Alignment.}
Preliminary experiments suggest that relying solely on the generated criteria leads to inconsistent scoring behavior in certain diagram types. This indicates that the judge VLM requires more specific guidance, particularly on how human experts interpret and apply the criteria during scoring. Therefore, we align the generated criteria with human preferences using the development set, so that the resulting criteria can better capture how human experts apply them in practice. By comparing reference diagrams against baseline generations (detailed in Section~\ref{sec:experiments}) under each criterion, we elicit human preferences and employ Gemini 3.1 Pro to distill them into unified preference statements. These statements are then appended to their corresponding criteria as additional scoring instructions.

\paragraph{Reference-Based Scoring.}
Since absolute scoring can be subjective and prone to scoring drift, we instead adopt a reference-based scoring strategy by providing the original human-created diagram $D$ as a reference anchor. To mitigate potential bias toward the reference diagram, the generated and reference diagrams are anonymized and presented in a randomized order alongside the type-specific criteria $\mathcal{C}_\tau$:
\begin{equation}
    \left(\{s_{\hat{D},c}\}_{c \in \mathcal{C}_\tau},
    \{s_{D,c}\}_{c \in \mathcal{C}_\tau}\right)
    = \operatorname{JudgeVLM}\!\left(P,C,D,\hat{D},\mathcal{C}_\tau\right).
    \label{eq:blind-scoring}
\end{equation}

Based on the two-level weights generated in the initial stage, the aggregated absolute score $S(\cdot)$ (where $\cdot$ represents either $\hat{D}$ or $D$) is computed by aggregating its corresponding criterion-level scores $s_{\cdot, c}$ (i.e., either $s_{\hat{D},c}$ or $s_{D,c}$) as:
\begin{equation}
    S(\cdot) = \sum_{d \in \mathcal{D}} W_{d,\tau}
    \sum_{c \in \mathcal{C}_{d,\tau}} w_c \cdot s_{\cdot,c},
    \label{eq:weighted-score}
\end{equation}
where $\mathcal{D}=\{\text{Structure},\text{Content},\text{Aesthetics}\}$. 
Preliminary experiments reveal that the absolute score of the reference diagram fluctuates depending on the quality of the generated diagram it is paired with. To provide a stable evaluation anchor and ensure comparability, the final score of the generated diagram is computed as a relative ratio:
\begin{equation}
    S_{\mathrm{final}}(\hat{D}) =
    \frac{S(\hat{D})}{S(\hat{D}) + S(D)}.
    \label{eq:relative-score}
\end{equation}

Accordingly, $S_{\mathrm{final}}(\hat{D})=0.5$ indicates parity with the reference diagram, while values below and above 0.5 indicate inferior and superior performance relative to the reference, respectively.

\section{Experiments}
\label{sec:experiments}

\subsection{Experimental Setup}
\label{sec:exp-setup}

\paragraph{Baselines and Evaluation Protocol.}
We evaluate four representative scientific diagram generation agents on \benchname{}: Vanilla, PaperBanana~\citep{zhu2026paperbanana}, AutoFigure~\citep{zhu2026autofigure}, and Paper2Figure~\citep{liu2025paper2any}. Vanilla is a simple two-stage baseline in which a VLM first converts the parsed paper into a generation prompt, and then a multimodal generative model generates the diagram. PaperBanana, AutoFigure, and Paper2Figure employ more complex multi-stage pipelines with explicit planning. More details of these agents are introduced in Section~\ref{related_work:sci_diag_gen}. Despite their different pipeline designs, all these agents rely on two major backbone components: a VLM to understand the paper and a multimodal generative model to generate the diagram. Unless otherwise specified, we use GPT-5.4~\citep{openai2026gpt54} as the default VLM and Gemini 3 Pro Image~\citep{google2025gemini3proimage} as the default multimodal generative model for all agents. We adopt  Gemini 3.1 Pro~\citep{google2026gemini31pro} as the default judge VLM. The temperature of all models is set to 0 to minimize stochastic variation.

\paragraph{Human Preference Annotation and Evaluation.}
To assess whether the proposed evaluation framework aligns with human preferences, we conduct human preference annotations based on the test set. For each instance, we generate a diagram using the Vanilla agent and pair it with the corresponding reference diagram for human comparison, resulting in a human preference dataset termed \benchname{}-VanillaPref. Three annotators, all Ph.D. students, independently assess these pairs using 5-point preference labels: strong reference-preferred, slight reference-preferred, unable to judge, slight generated-preferred, and strong generated-preferred. To resolve disagreements among annotators, we further design a set of aggregation rules to derive the final preference labels. After aggregation, we retain 105 high-agreement pairs, comprising 24 strong reference-preferred, 46 slight reference-preferred, 29 slight generated-preferred, and 6 strong generated-preferred pairs. The remaining 5 pairs are excluded from subsequent evaluation due to conflicting directional judgments among annotators. The annotation guidelines and aggregation rules are provided in Appendix~\ref{sec:appendix-human-annotations}.

We evaluate human preference alignment from two perspectives: \emph{directional agreement} and \emph{strength alignment}. For directional agreement, we use Directional Agreement Rate (DAR), which measures whether the preference direction indicated by the automatic relative score is consistent with the human judgment, using 0.5 as the direction threshold. For strength alignment, we measure the correspondence between automatic evaluation scores and human preference strength using Kendall's $\tau_b$~\citep{kendall1938new} and Spearman's $\rho$~\citep{spearman1904proof}. Kendall's $\tau_b$ evaluates ordinal concordance while explicitly accounting for ties, whereas Spearman's $\rho$ measures the monotonic association between the two rankings.

\subsection{Main Results}
\label{sec:main-results}

\paragraph{Results across Diagram Types.}
To examine the overall performance of different scientific diagram generation agents across diagram types, we report the relative score of each agent on every diagram type as well as its overall score across all types. As shown in Table~\ref{tab:main-results-types}, although PaperBanana achieves the highest overall score (0.5312), the relative performance of these agents varies considerably across diagram types, and no single pipeline design proves uniformly optimal. Generally, current agents perform relatively well on diagram types that emphasize high-level information organization and abstract conceptual relationships. For instance, top-performing agents even surpass the human-parity threshold (0.5) on types such as Graphical Abstracts and Conceptual Frameworks. By contrast, for types such as Illustrative Case and Lineage Dendrogram, which demand concrete and precise structures, current agents still fall short.

\begin{table}[!t]
\caption{Results across different diagram types on \benchname{}.}
\label{tab:main-results-types}
\centering
\vspace{\baselineskip}
\resizebox{\textwidth}{!}{%
\begin{tabular}{lcccccccccccc}
\toprule
Agent & TL & LD & PE & AF & CSA & CF & TR & GA & ESD & IC & MD & Overall \\
\midrule
Vanilla & 0.4406 & 0.4168 & 0.5626 & 0.4670 & \textbf{0.4947} & \textbf{0.5820} & 0.5094 & \textbf{0.6228} & \textbf{0.5258} & 0.3740 & 0.4566 & 0.4957 \\
PaperBanana & \textbf{0.5272} & \textbf{0.4917} & 0.5975 & \textbf{0.5665} & 0.4926 & 0.5703 & \textbf{0.5361} & 0.6195 & 0.4610 & 0.4824 & \textbf{0.4985} & \textbf{0.5312} \\
Paper2Figure & 0.4714 & 0.4514 & \textbf{0.6060} & 0.4314 & 0.3337 & 0.4594 & 0.4057 & 0.6034 & 0.4555 & \textbf{0.4865} & 0.3241 & 0.4571 \\
AutoFigure & 0.2172 & 0.1162 & 0.1189 & 0.2132 & 0.2222 & 0.1919 & 0.2905 & 0.2956 & 0.1604 & 0.1476 & 0.2137 & 0.1989 \\
\bottomrule
\end{tabular}}
\end{table}

\paragraph{Results across Evaluation Dimensions.}
To examine the performance of different agents across evaluation dimensions, we report their scores on Structure, Content, and Aesthetics, together with the corresponding sub-dimensions, in Table~\ref{tab:main-results-dimensions}. As shown in the table, PaperBanana achieves the highest scores across all dimensions and is the only agent that consistently surpasses the human-parity threshold. Comparing top-level dimensions, the Content scores are consistently lower than the Structure scores across all agents, indicating that accurately representing scientific content remains more challenging than organizing the overall diagram structure for existing agents. Moreover, for Vanilla, PaperBanana, and Paper2Figure, Aesthetics consistently receives the highest top-level score, suggesting that current agents are relatively effective at producing visually appealing diagrams. A similar pattern can be observed within the Structure and Content dimensions. For Vanilla, PaperBanana, and Paper2Figure, VOL outperforms LA, while VER outperforms CA. This suggests that current agents are generally stronger at visually organizing and presenting information than at preserving rigorous logical and semantic accuracy.

\begin{table}[!t]
\caption{Results across different evaluation dimensions on \benchname{}.}
\label{tab:main-results-dimensions}
\centering
\vspace{\baselineskip}
\resizebox{0.75\textwidth}{!}{
\begin{tabular}{lccccccc}
\toprule
\multirow{2}{*}{Method} & \multicolumn{3}{c}{Structure} & \multicolumn{3}{c}{Content} & \multirow{2}{*}{Aesthetics} \\
\cmidrule(lr){2-4} \cmidrule(lr){5-7}
 & VOL & LA & Weighted & VER & CA & Weighted & \\
\midrule
Vanilla & 0.5083 & 0.4756 & 0.4909 & 0.5057 & 0.4650 & 0.4829 & 0.5393 \\
PaperBanana & \textbf{0.5412} & \textbf{0.5259} & \textbf{0.5326} & \textbf{0.5318} & \textbf{0.5158} & \textbf{0.5224} & \textbf{0.5544} \\
Paper2Figure & 0.4881 & 0.4092 & 0.4495 & 0.4916 & 0.3991 & 0.4433 & 0.5108 \\
AutoFigure & 0.2318 & 0.1892 & 0.2106 & 0.1549 & 0.1943 & 0.1808 & 0.2020 \\
\bottomrule
\end{tabular}}
\end{table}

\paragraph{AutoFigure Failure Analysis.}
As AutoFigure exhibits substantially lower performance, we further analyze its generated diagrams to identify the underlying failure patterns. As shown in Tables~\ref{tab:main-results-types} and~\ref{tab:main-results-dimensions}, its performance is particularly low on LD (0.1162) and PE (0.1189), as well as on the VER dimension (0.1549), suggesting difficulties in handling complex topological structures, continuous transformations, and faithful visual-element representation. Further analysis of the generated diagrams reveals several failure patterns, including replacing visual elements with lengthy text blocks, using overly generic geometric primitives or literal placeholders, and producing unclear connections between diagram components. Representative cases are provided in Appendix~\ref{sec:appendix-qualitative-case}. Such difficulties may partly reflect the challenge of encoding complex visual structures through its SVG-based intermediate representation.

\subsection{Sensitivity Analysis of Judge Models}
\label{sec:evaluation-reliability}

Since the capability and scoring behavior of the judge model can substantially
affect automatic scientific diagram evaluation, we first examine the sensitivity
of evaluation results to different judge models. Specifically, in addition to
the default Gemini 3.1 Pro, we use GPT-5.4 as an alternative judge. 

\paragraph{Main Results with Different Judge Models.}

\begin{figure}[!t]
\begin{center}
\includegraphics[width=\textwidth]{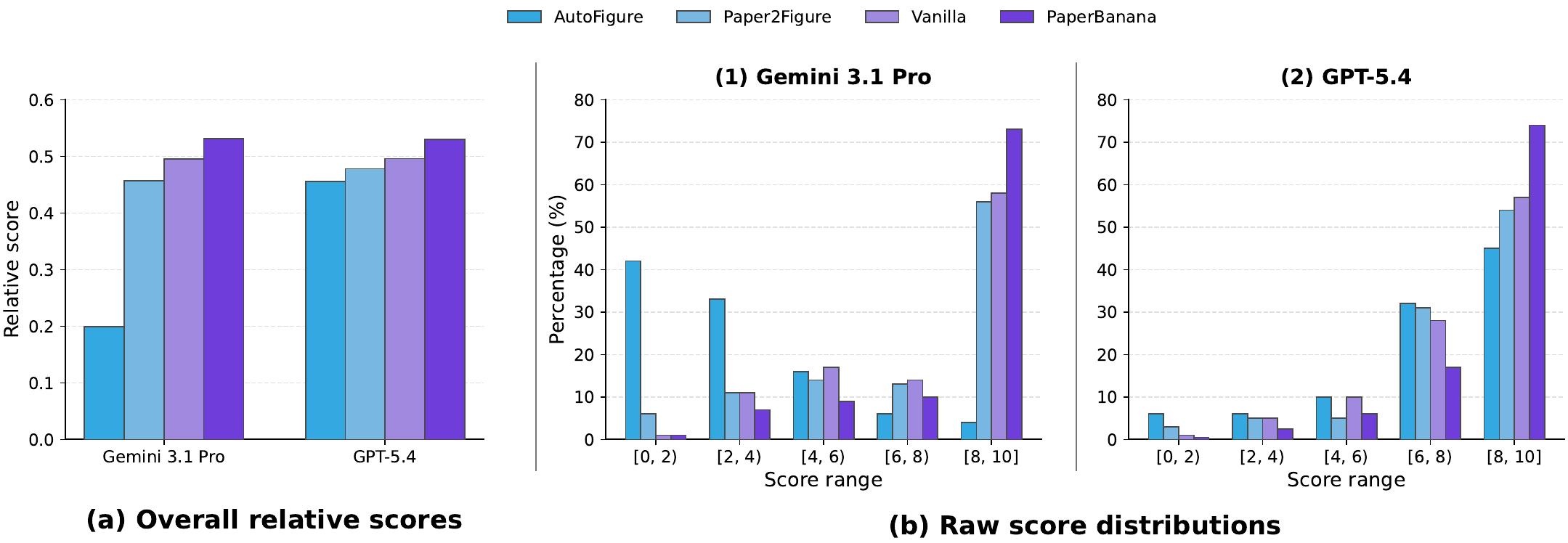}
\caption{Sensitivity of evaluation results to different judge models.
(a) Overall relative scores of different generation agents.
(b) Raw score distributions produced by Gemini 3.1 Pro and GPT-5.4.}
\label{fig:judge-score-distribution}
\end{center}
\end{figure}

To examine the effect of the judge model on automatic evaluation, we
re-evaluate the outputs of all generation agents on the
test set using GPT-5.4. As shown in
Figure~\ref{fig:judge-score-distribution}(a), although the two judges preserve
the same ordering among generation agents, the score gaps among them are substantially reduced under GPT-5.4. Among the four agents, AutoFigure shows the largest score increase under GPT-5.4, while the scores of Vanilla and PaperBanana remain relatively stable, and Paper2Figure shows a little increase.

Figure~\ref{fig:judge-score-distribution}(b) further compares the raw-score
distributions produced by the two judges. GPT-5.4 assigns substantially fewer
scores in the low-score ranges and concentrates more scores in the high-score
ranges, while Gemini 3.1 Pro produces more differentiated distributions across
different agents. For example, when evaluating AutoFigure, 41.7\% of the scores assigned by Gemini 3.1 Pro fall within the $[0,2)$ interval, compared with only 5.9\% under GPT-5.4.

To further analyze this discrepancy, we check the diagrams generated by
AutoFigure. Among the 110 generated diagrams, 98 contain severe defects, such as
missing glyphs and overlapping text. However, GPT-5.4 fails to
identify these defects in 52 cases, including 39 cases in which it positively
rationalizes the defects and assigns high scores. In the remaining 46 cases, although the defects are explicitly identified, GPT-5.4 does not penalize them as strongly as prescribed by the rubric.

\begin{table}[!t]
\caption{Human preference alignment evaluation on \benchname{}-VanillaPref. Ref. and Gen. denote reference-preferred and generated-preferred, respectively. $\Delta$ denotes the judge score minus 0.5, which indicates the direction and strength of the judge preference.}
\label{tab:judge-human-strata}
\centering
\vspace{\baselineskip}
\footnotesize
\begin{tabular}{lcccccc}
\toprule
Human stratum & GPT & $\Delta$ GPT & Gemini & $\Delta$ Gemini & GPT DAR & Gemini DAR \\
\midrule
Strong Ref. & 0.4690 & -0.0310 & 0.3806 & -0.1194 & 0.5417 & 0.8750 \\
Slight Ref. & 0.4882 & -0.0118 & 0.4357 & -0.0643 & 0.4783 & 0.7826 \\
Slight Gen. & 0.5210 & +0.0210 & 0.6249 & +0.1249 & 0.6552 & 0.9310 \\
Strong Gen. & 0.5511 & +0.0511 & 0.6838 & +0.1838 & 1.0000 & 1.0000 \\
\bottomrule
\end{tabular}
\end{table}

\paragraph{Evaluation of Human Preference Alignment across Judge Models.}

Given the substantial discrepancies between the two judge models, we further assess how well their evaluation results align with human preferences on \benchname{}-VanillaPref. This analysis also allows us to examine whether the different scoring behaviors persist on Vanilla outputs, which contain fewer severe defects than AutoFigure outputs. As shown in Table~\ref{tab:judge-human-strata}, the judge scores given by GPT-5.4 exhibit substantially smaller deviations from the 0.5 parity threshold (ranging from $-0.031$ to $+0.051$) compared to those given by Gemini 3.1 Pro (ranging from $-0.119$ to $+0.184$) across the four human-preference strata, indicating a strong tendency of GPT-5.4 to compress relative scores toward 0.5. This compression makes its evaluations less consistent with human preference strength. Under the complete evaluation framework (Tables~\ref{tab:judge-framework-validity} and~\ref{tab:appendix-judge-framework-validity}), Gemini 3.1 Pro shows substantially higher correlations with human judgments than GPT-5.4, with Kendall's $\tau_b$ of 0.594 vs.\ 0.228 and Spearman's $\rho$ of 0.733 vs.\ 0.293. GPT-5.4 also exhibits a notably lower DAR with human judgments, particularly on reference-preferred instances (Table~\ref{tab:judge-human-strata}).

\subsection{Ablation Study}
\label{sec:ablation-study}

To investigate how Type-Specific Criteria (TSC) and Human Preference Alignment (HPA) affect the evaluation results, we compare four configurations: ours, w.o. HPA, w.o. TSC, and w.o. HPA\&TSC. w.o. TSC replaces the type-specific criteria with a unified set of general criteria shared across all diagram types, while w.o. HPA removes the preference statements from the evaluation criteria while retaining the type-specific criteria. w.o. HPA\&TSC removes both components. The results on \benchname{}-VanillaPref are in Table~\ref{tab:judge-framework-validity}. We use the Directional Agreement Rate (DAR with 0.5 as the decision threshold) as the primary metric, together with correlation metrics such as Kendall's $\tau_b$ to measure agreement with human preference annotation.

\begin{table}[!t]
\caption{Ablation results for preference alignment evaluation on
\benchname{}-VanillaPref.}
\label{tab:judge-framework-validity}
\centering
\vspace{\baselineskip}
\footnotesize
\begin{tabular*}{0.9\textwidth}{@{\extracolsep{\fill}}ccccccc@{}}
\toprule
\multicolumn{2}{c}{Experimental setup} & \multicolumn{3}{c}{Directional Agreement Rate} & \multicolumn{2}{c}{Strength Alignment} \\
\cmidrule(lr){1-2} \cmidrule(lr){3-5} \cmidrule(lr){6-7}
TSC & HPA & Overall & Gen.-pref. & Ref.-pref. & Kendall's $\tau_b$ & Spearman's $\rho$ \\
\midrule
-- & -- & 0.6952 & 0.9143 & 0.5857 & 0.4787 & 0.5979 \\
-- & \checkmark & 0.7048 & 0.9429 & 0.5857 & 0.2085 & 0.2746 \\
\checkmark & -- & 0.7619 & 0.9429 & 0.6714 & 0.4974 & 0.6205 \\
\checkmark & \checkmark & \textbf{0.8571} & 0.9429 & \textbf{0.8143} & \textbf{0.5943} & \textbf{0.7329} \\
\bottomrule
\end{tabular*}
\end{table}

The ablation results demonstrate that both TSC and HPA contribute to human-aligned evaluation, while their combination yields the largest improvement. Removing HPA decreases the overall DAR from 0.857 to 0.762 and the reference-preferred DAR from 0.814 to 0.671, whereas removing TSC leads to a larger drop to 0.705 and 0.586, respectively. Notably, HPA alone provides only marginal gains over the configuration without either component, while its effect becomes much stronger when combined with TSC. This suggests that TSC provides a more informative type-specific evaluation basis, upon which HPA can further calibrate the evaluation toward human preferences.

\subsection{Stability Analysis}
\label{sec:repeated-scoring-stability}

To assess the stability of our evaluation framework, we perform two additional scoring runs on the test set, yielding three independent runs in total. We compare four configurations: ours, w.o. HPA, w.o. TSC, and w.o. HPA\&TSC. Across the four compared configurations, we evaluate stability using three groups of metrics: 1) consistency metrics, including Intraclass Correlation Coefficient (ICC), between-instance, and within-instance variance; 2) categorical stability, measured by direction flip rate and criterion flip rate; and 3) criterion entropy, which quantifies the dispersion of criterion-level scores across runs. For flip-rate metrics, we discretize evaluation outcomes into three directions (generated-preferred, tied, and reference-preferred), using a threshold of 0 for criterion-level score differences and 0.5 for instance-level relative scores. The criterion flip rate is computed as the unweighted average over individual criteria, whereas the direction flip rate is computed from the instance-level relative scores defined in Eq.~(\ref{eq:relative-score}).

As shown in Table~\ref{tab:repeated-scoring-stability}, all four configurations exhibit relatively high reproducibility, with ICC values ranging from 0.8613 to 0.8890 and within-instance variance remaining substantially smaller than between-instance variance. The direction flip rates are also relatively similar across configurations, ranging from 0.1212 to 0.1333. Notably, no single configuration consistently achieves the best value across all stability metrics. For example, the configuration without TSC or HPA achieves the highest ICC (0.8890), whereas configurations with HPA obtain a slightly lower direction flip rate (0.1212). These results suggest that the evaluation framework is generally reproducible under repeated scoring, while the effects of individual components on aggregate stability remain relatively small. A more detailed type-level stability analysis is provided in Appendix~\ref{sec:appendix-type-level-stability}.

\begin{table}[!t]
\caption{Stability over three independent scoring runs. Between var. and within var. denote between-instance and within-instance variance.}
\label{tab:repeated-scoring-stability}
\centering
\vspace{\baselineskip}
\small
\setlength{\tabcolsep}{4pt}
\begin{tabular}{cccccccc}
\toprule
\multicolumn{2}{c}{Experimental setup} & \multicolumn{6}{c}{Stability metrics} \\
\cmidrule(lr){1-2} \cmidrule(lr){3-8}
TSC & HPA & ICC & Between var. & Within var. & Direction flip & Criterion flip & Criterion entropy \\
\midrule
-- & -- & 0.8890 & 0.0119 & 0.00149 & 0.1333 & 0.1958 & 0.1727 \\
-- & \checkmark & 0.8718 & 0.0113 & 0.00166 & 0.1212 & 0.1939 & 0.1708 \\
\checkmark & -- & 0.8613 & 0.0141 & 0.00228 & 0.1333 & 0.1883 & 0.1664 \\
\checkmark & \checkmark & 0.8714 & 0.0177 & 0.00261 & 0.1212 & 0.2030 & 0.1797 \\
\bottomrule
\end{tabular}
\end{table}

\section{Conclusion}
In this work, we introduce \benchname{}, a type-specific benchmark for human-aligned evaluation of scientific diagram generation. \benchname{} provides a systematically categorized dataset of 165 diagrams spanning 11 diagram types and 10 scientific domains, together with an evaluation framework that combines type-specific criteria with preference-guided calibration to better reflect human judgments. Extensive experiments on representative agents demonstrate the importance of accounting for diagram-type differences and human evaluation preferences.

\bibliography{references}

@article{xu2026ai4science,
author = {Xu, Yongjun and Shao, Zezhi and Liu, Xin and Li, Yuquan and An, Zhulin and Fu, Chenguang and Huang, Changping and Zhang, Lin and Gao, Feng and Li, Jian and others},
title = {{AI} for science: Progress, challenges, and perspectives},
journal = {The Innovation},
year = {2026},
doi = {10.1016/j.xinn.2026.101530},
url = {https://www.cell.com/the-innovation/fulltext/S2666-6758(26)00277-8}
}

@misc{google2025gemini3proimage,
author = {{Google}},
title = {{Gemini 3 Pro Image}},
year = {2025},
howpublished = {\url{https://ai.google.dev/gemini-api/docs/models/gemini-3-pro-image}},
note = {Accessed: 2026-06-27}
}

@article{kendall1938new,
author = {Kendall, M. G.},
title = {A New Measure of Rank Correlation},
journal = {Biometrika},
volume = {30},
number = {1--2},
pages = {81--93},
year = {1938},
doi = {10.1093/biomet/30.1-2.81}
}

@article{spearman1904proof,
author = {Spearman, C.},
title = {The Proof and Measurement of Association between Two Things},
journal = {The American Journal of Psychology},
volume = {15},
number = {1},
pages = {72--101},
year = {1904},
doi = {10.2307/1412159}
}

@misc{openai2026gptimage2,
author = {{OpenAI}},
title = {{GPT Image 2}: State-of-the-art image generation model},
year = {2026},
howpublished = {\url{https://developers.openai.com/api/docs/models/gpt-image-2}},
note = {Accessed: 2026-06-27}
}

@misc{alibaba2026qwen35plus,
author = {{Alibaba Cloud}},
title = {{Qwen3.5-Plus}},
year = {2026},
howpublished = {\url{https://www.alibabacloud.com/help/en/model-studio/text-generation-model/}},
note = {Accessed: 2026-08-19}
}

@misc{bytedance2026seed2,
author = {{ByteDance Seed Team}},
title = {{Seed2.0} Model Card: Towards Intelligence Frontier for Real-World Complexity},
year = {2026},
howpublished = {\url{https://github.com/ByteDance-Seed/Seed2.0}},
note = {Accessed: 2026-06-27}
}

@misc{openai2026gpt54,
author = {{OpenAI}},
title = {{GPT-5.4}},
year = {2026},
howpublished = {\url{https://developers.openai.com/api/docs/models/gpt-5.4}},
note = {Accessed: 2026-06-27}
}

@misc{google2026gemini31pro,
author = {{Google}},
title = {{Gemini 3.1 Pro}},
year = {2026},
howpublished = {\url{https://ai.google.dev/gemini-api/docs/models/gemini-3.1-pro}},
note = {Accessed: 2026-06-27}
}

@misc{bytedance2026seedream5,
author = {{ByteDance Seed Team}},
title = {{Seedream 5}},
year = {2026},
howpublished = {\url{https://seed.bytedance.com/tech/seedream5}},
note = {Accessed: 2026-06-27}
}

@article{huang2026scifig,
title={{SciFig}: Towards Automating Scientific Figure Generation},
author={Huang, Siyuan and Gao, Yutong and Bai, Juyang and Zhou, Yifan and Yin, Zi and Liu, Xinxin and Chellappa, Rama and Lau, Chun Pong and Nag, Sayan and Peng, Cheng and Pramanick, Shraman},
journal={arXiv preprint arXiv:2601.04390},
year={2026}
}

@article{zhu2026paperbanana,
title={{PaperBanana}: Automating Academic Illustration for {AI} Scientists},
author={Zhu, Dawei and Meng, Rui and Song, Yale and Wei, Xiyu and Li, Sujian and Pfister, Tomas and Yoon, Jinsung},
journal={arXiv preprint arXiv:2601.23265},
year={2026}
}

@inproceedings{zhu2026autofigure,
title={AutoFigure: Generating and Refining Publication-Ready Scientific Illustrations},
author={Minjun Zhu and Zhen Lin and Yixuan Weng and Panzhong Lu and Qiujie Xie and Yifan Wei and Sifan Liu and QiYao Sun and Yue Zhang},
booktitle={The Fourteenth International Conference on Learning Representations},
year={2026},
url={https://openreview.net/forum?id=5N3z9JQJKq}
}

@misc{alibaba2026qwenimage20pro,
author = {{Alibaba Cloud}},
title = {{Qwen-Image-2.0-Pro}},
year = {2026},
howpublished = {\url{https://www.alibabacloud.com/help/en/model-studio/qwen-image-api}},
note = {Accessed: 2026-08-19}
}

@inproceedings{mondal2024scidoc2diagrammer,
title={{SciDoc2Diagrammer-MAF}: Towards Generation of Scientific Diagrams from Documents guided by Multi-Aspect Feedback Refinement},
author={Mondal, Ishani and Li, Zongxia and Hou, Yufang and Natarajan, Anandhavelu and Garimella, Aparna and Boyd-Graber, Jordan},
booktitle={Findings of the Association for Computational Linguistics: EMNLP 2024},
pages={13342--13375},
address={Miami, Florida, USA},
publisher={Association for Computational Linguistics},
year={2024}
}

@inproceedings{li2026easier,
title={Easier Painting Than Thinking: Can Text-to-Image Models Set the Stage, but Not Direct the Play?},
author={Li, Ouxiang and Wang, Yuan and Hu, Xinting and Huang, Huijuan and Chen, Rui and Ou, Jiarong and Tao, Xin and Wan, Pengfei and Qi, Xiaojuan and Feng, Fuli},
booktitle={The Fourteenth International Conference on Learning Representations},
year={2026}
}

@inproceedings{hessel2021clipscore,
title={{CLIPScore}: A Reference-Free Evaluation Metric for Image Captioning},
author={Hessel, Jack and Holtzman, Ari and Forbes, Maxwell and Le Bras, Ronan and Choi, Yejin},
booktitle={Proceedings of the 2021 Conference on Empirical Methods in Natural Language Processing (EMNLP)},
pages={7514--7528},
address={Online and Punta Cana, Dominican Republic},
publisher={Association for Computational Linguistics},
year={2021}
}

@inproceedings{heusel2017gans,
title={{GANs} Trained by a Two Time-Scale Update Rule Converge to a Local Nash Equilibrium},
author={Heusel, Martin and Ramsauer, Hubert and Unterthiner, Thomas and Nessler, Bernhard and Hochreiter, Sepp},
booktitle={Advances in Neural Information Processing Systems},
volume={30},
year={2017}
}

@inproceedings{jayasumana2024rethinking,
title={Rethinking {FID}: Towards a Better Evaluation Metric for Image Generation},
author={Jayasumana, Sadeep and Ramalingam, Srikumar and Veit, Andreas and Glasner, Daniel and Chakrabarti, Ayan and Kumar, Sanjiv},
booktitle={Proceedings of the IEEE/CVF Conference on Computer Vision and Pattern Recognition (CVPR)},
pages={9307--9315},
year={2024}
}

@inproceedings{luo2025mmmg,
title={{MMMG}: A Massive, Multidisciplinary, Multi-Tier Generation Benchmark for Text-to-Image Reasoning},
author={Luo, Yuxuan and Yuan, Yuhui and Chen, Junwen and Cai, Haonan and Yue, Ziyi and Yang, Yuwei and Daha, Fatima Zohra and Li, Ji and Lian, Zhouhui},
booktitle={Advances in Neural Information Processing Systems 38 (NeurIPS 2025) Datasets and Benchmarks Track},
year={2025}
}

@article{chang2026fepbench,
title={Faithful, Enriched, and Precise: Benchmarking Natural-Science Illustration Generation by {T2I} models},
author={Chang, Yifan and Ai, Jiaxin and Sun, Jianwen and Pu, Yuandong and Luo, Siqi and Zhao, Liangliang and Ren, Yuchen and Liu, Minghao and Yu, Yunfei and Qiao, Yu and Zhang, Kaipeng and Liu, Yihao},
journal={arXiv preprint arXiv:2606.05949},
year={2026}
}

@article{zhao2026kvbench,
title={Knowledge Visualization: A Benchmark and Method for Knowledge-Intensive Text-to-Image Generation},
author={Zhao, Ran and Jin, Sheng and Wu, Size and Liao, Kang and Gong, Zerui and Guo, Zujin and Xiao, Yang and Li, Wei},
journal={arXiv preprint arXiv:2604.22302},
year={2026}
}

@article{li2026bizgeneval,
title={{BizGenEval}: A Systematic Benchmark for Commercial Visual Content Generation},
author={Li, Yan and Zeng, Zezi and Zhou, Ziwei and Gao, Xin and Tian, Muzhao and Yang, Yifan and Cheng, Mingxi and Dai, Qi and Yang, Yuqing and Qiu, Lili and Wang, Zhendong and Yang, Zhengyuan and Yang, Xue and Wang, Lijuan and Li, Ji and Luo, Chong},
journal={arXiv preprint arXiv:2603.25732},
year={2026}
}

@inproceedings{sun2026t2ireasonbench,
title={{T2I-ReasonBench}: Benchmarking Reasoning-Informed Text-to-Image Generation},
author={Sun, Kaiyue and Fang, Rongyao and Duan, Chengqi and Liu, Xian and Li, Aoxue and Liu, Xihui},
booktitle={Findings of the Association for Computational Linguistics: ACL 2026},
pages={8919--8944},
address={San Diego, California, United States},
publisher={Association for Computational Linguistics},
year={2026}
}

@inproceedings{pang2025paper2poster,
title={{Paper2Poster}: Towards Multimodal Poster Automation from Scientific Papers},
author={Pang, Wei and Lin, Kevin Qinghong and Jian, Xiangru and He, Xi and Torr, Philip},
booktitle={Advances in Neural Information Processing Systems 38 (NeurIPS 2025) Datasets and Benchmarks Track},
year={2025}
}

@article{chang2025sridbench,
title={{SridBench}: Benchmark of Scientific Research Illustration Drawing of Image Generation Model},
author={Chang, Yifan and Feng, Yukang and Sun, Jianwen and Ai, Jiaxin and Li, Chuanhao and Zhou, S. Kevin and Zhang, Kaipeng},
journal={arXiv preprint arXiv:2505.22126},
year={2025}
}

@inproceedings{belouadi2025tikzero,
title={{TikZero}: Zero-Shot Text-Guided Graphics Program Synthesis},
author={Belouadi, Jonas and Ilg, Eddy and Keuper, Margret and Tanaka, Hideki and Utiyama, Masao and Dabre, Raj and Eger, Steffen and Ponzetto, Simone Paolo},
booktitle={Proceedings of the IEEE/CVF International Conference on Computer Vision (ICCV)},
pages={17793--17806},
year={2025}
}

@inproceedings{ge2025autopresent,
title={{AutoPresent}: Designing Structured Visuals from Scratch},
author={Ge, Jiaxin and Wang, Zora Zhiruo and Zhou, Xuhui and Peng, Yi-Hao and Subramanian, Sanjay and Tan, Qinyue and Sap, Maarten and Suhr, Alane and Fried, Daniel and Neubig, Graham and Darrell, Trevor},
booktitle={Proceedings of the IEEE/CVF Conference on Computer Vision and Pattern Recognition (CVPR)},
pages={2902--2911},
year={2025}
}

@inproceedings{du2026deepresearchbench,
title={{DeepResearch Bench}: A Comprehensive Benchmark for Deep Research Agents},
author={Du, Mingxuan and Xu, Benfeng and Zhu, Chiwei and Zhang, Licheng and Wang, Xiaorui and Mao, Zhendong},
booktitle={The Fourteenth International Conference on Learning Representations},
year={2026},
url={https://openreview.net/forum?id=hQ0K2Hhq7H}
}

@inproceedings{zheng2025pptagent,
title={{PPTAgent}: Generating and Evaluating Presentations Beyond Text-to-Slides},
author={Zheng, Hao and Guan, Xinyan and Kong, Hao and Zhang, Wenkai and Zheng, Jia and Zhou, Weixiang and Lin, Hongyu and Lu, Yaojie and Han, Xianpei and Sun, Le},
booktitle={Proceedings of the 2025 Conference on Empirical Methods in Natural Language Processing (EMNLP)},
pages={14402--14418},
address={Suzhou, China},
publisher={Association for Computational Linguistics},
year={2025}
}

@inproceedings{liang2025diagrameval,
title={{DiagramEval}: Evaluating {LLM}-Generated Diagrams via Graphs},
author={Liang, Chumeng and You, Jiaxuan},
booktitle={Proceedings of the 2025 Conference on Empirical Methods in Natural Language Processing (EMNLP)},
pages={12678--12690},
address={Suzhou, China},
publisher={Association for Computational Linguistics},
url={https://aclanthology.org/2025.emnlp-main.640/},
doi={10.18653/v1/2025.emnlp-main.640},
year={2025}
}

@article{dhote2023figuresurvey,
title={A Survey on Figure Classification Techniques in Scientific Documents},
author={Dhote, Anurag and Javed, Mohammed and Doermann, David S},
journal={arXiv preprint arXiv:2307.05694},
year={2023}
}

@inproceedings{jobin2019docfigure,
title={{DocFigure}: A Dataset for Scientific Document Figure Classification},
author={Jobin, K. V. and Mondal, A. and Jawahar, C. V.},
booktitle={2019 International Conference on Document Analysis and Recognition Workshops (ICDARW)},
pages={74--79},
organization={IEEE},
year={2019}
}

@inproceedings{karishma2023aclfig,
title={{ACL-FIG}: A Dataset for Scientific Figure Classification},
author={Karishma, Z. and Rohatgi, S. and Puranik, K. S. and Wu, J. and Giles, C. L.},
booktitle={Proceedings of the Workshop on Scientific Document Understanding (SDU 2023)},
year={2023}
}

@article{niu2025mineru25,
title={{MinerU2.5}: A Decoupled Vision-Language Model for Efficient High-Resolution Document Parsing},
author={Niu, Junbo and Liu, Zheng and Gu, Zhuangcheng and Wang, Bin and Ouyang, Linke and Zhao, Zhiyuan and Chu, Tao and He, Tianyao and Wu, Fan and Zhang, Qintong and others},
journal={arXiv preprint arXiv:2509.22186},
year={2025}
}

@misc{liu2025paper2any,
title={{Paper2Any}: Turn Paper/Text/Topic into Editable Research Figures, Technical Route Diagrams, and Presentation Slides},
author={Liu, Z. and Guo, Z. and Su, Z. and Huang, L. and Yang, Y. and Han, Z. and Pan, Z. and Zhang, W.},
year={2025},
howpublished={GitHub repository},
url={https://github.com/OpenDCAI/Paper2Any},
note={Accessed: 2026-08-26}
}

@article{liu2025lignin,
  author  = {Liu, Ruo-Ying and Li, Bing-Zhi and Yuan, Ying-Jin and Liu, Zhi-Hua},
  title   = {Multiscale metabolic engineering in biological lignin valorization},
  journal = {The Innovation},
  volume  = {6},
  number  = {11},
  pages   = {100993},
  year    = {2025},
  doi     = {10.1016/j.xinn.2025.100993},
  url     = {https://doi.org/10.1016/j.xinn.2025.100993}
}

@article{chalouan2023cgps,
  author  = {Chalouan, A. and Gil, A. J. and Chabli, A. and Bargach, K. and Liemlahi, H. and El Kadiri, K. and Tendero-Salmeron, V. and Galindo-Zaldivar, J.},
  title   = {cGPS Record of Active Extension in Moroccan Meseta and Shortening in Atlasic Chains under the Eurasia-Nubia Convergence},
  journal = {Sensors},
  volume  = {23},
  pages   = {4846},
  year    = {2023},
  doi     = {10.3390/s23104846},
  url     = {https://doi.org/10.3390/s23104846}
}

@article{li2025haishao1,
  author  = {Li, Xiao-Ming and Chang, Liang and Zhou, Jian and Sun, Jili and Lu, Xingqiang and Han, Xingbo and Wang, Jianping and Lu, Hongliang and Li, Xing and Wang, Ning and Qiu, Yujia and Wu, Yirong},
  title   = {Haishao-1 satellite: Low-inclination orbit spaceborne synthetic aperture radar},
  journal = {The Innovation},
  volume  = {6},
  number  = {9},
  pages   = {100949},
  year    = {2025},
  doi     = {10.1016/j.xinn.2025.100949},
  url     = {https://doi.org/10.1016/j.xinn.2025.100949}
}

@article{danca2024mandelbrot,
  author  = {Danca, Marius-F.},
  title   = {Mandelbrot Set as a Particular Julia Set of Fractional Order, Equipotential Lines and External Rays of Mandelbrot and Julia Sets of Fractional Order},
  journal = {Fractal and Fractional},
  volume  = {8},
  number  = {1},
  pages   = {69},
  year    = {2024},
  doi     = {10.3390/fractalfract8010069},
  url     = {https://doi.org/10.3390/fractalfract8010069}
}

@article{anton2025intentas,
  author  = {Anton, O. and others},
  title   = {{INTENTAS} -- an entanglement-enhanced atomic sensor for microgravity},
  journal = {EPJ Quantum Technology},
  volume  = {12},
  pages   = {26},
  year    = {2025},
  doi     = {10.1140/epjqt/s40507-025-00330-9},
  url     = {https://doi.org/10.1140/epjqt/s40507-025-00330-9}
}

@article{wang2025untethered,
  author  = {Wang, Q. and others},
  title   = {Untethered miniature robots for minimally invasive thrombus treatment: From bench to clinical trials},
  journal = {The Innovation},
  volume  = {6},
  number  = {6},
  pages   = {100874},
  year    = {2025},
  doi     = {10.1016/j.xinn.2025.100874},
  url     = {https://doi.org/10.1016/j.xinn.2025.100874}
}

@inproceedings{shi2026judgeagent,
  title={{JudgeAgent}: Beyond Static Benchmarks for Knowledge-Driven and Dynamic {LLM} Evaluation},
  author={Shi, Zhichao and Jiang, Xuhui and Xu, Chengjin and Yao, Cangli and Ma, Shengjia and Shen, Yinghan and Li, Zixuan and Guo, Jian and Wang, Yuanzhuo},
  booktitle={Findings of the Association for Computational Linguistics: ACL 2026},
  year={2026},
  publisher={Association for Computational Linguistics},
  url={https://aclanthology.org/2026.findings-acl.634/}
}
\bibliographystyle{scidiagbench}

\appendix

\section{Dataset Construction Details: Filtering and Composition}
\label{sec:dataset-construction-details}

\subsection{Representative Examples of Excluded Diagram Categories}
\label{sec:appendix-excluded-categories}

\begin{figure}[H]
    \centering
    \begin{minipage}[t]{0.32\textwidth}
        \centering
        \includegraphics[width=\linewidth,height=0.20\textheight,keepaspectratio]{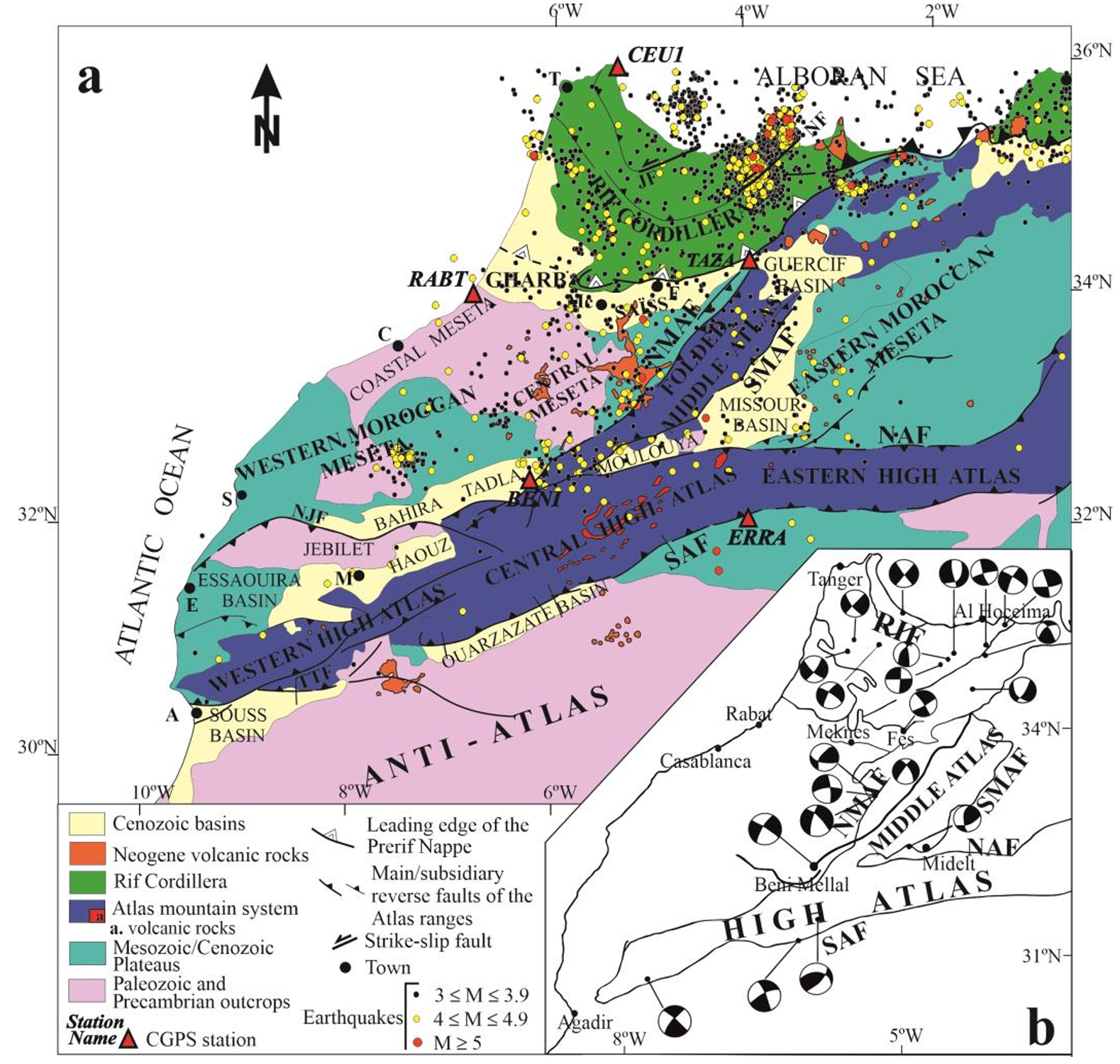}
        \par\smallskip\small\textbf{(a)} Data-driven map
    \end{minipage}\hfill
    \begin{minipage}[t]{0.32\textwidth}
        \centering
        \includegraphics[width=\linewidth,height=0.20\textheight,keepaspectratio]{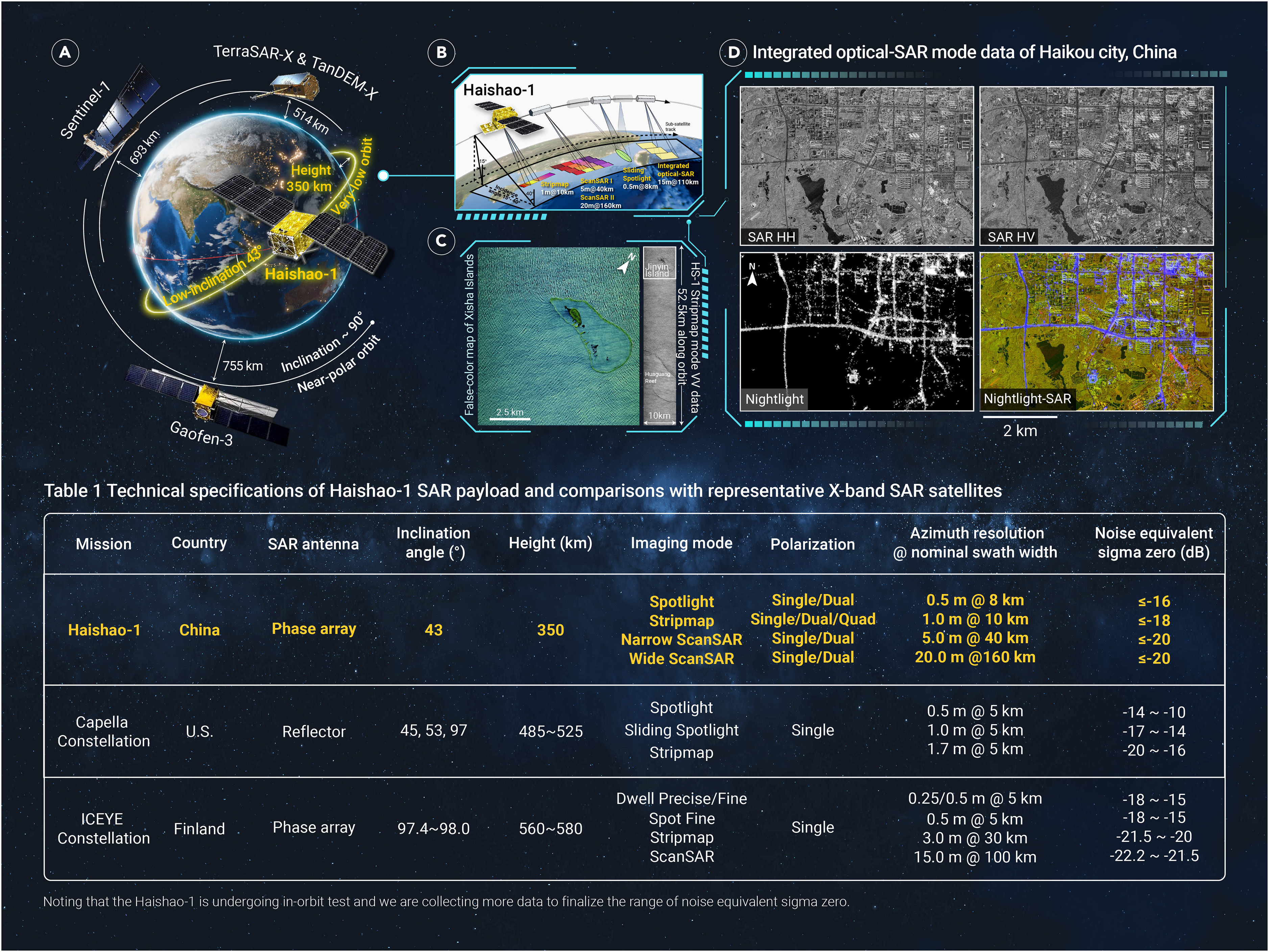}
        \par\smallskip\small\textbf{(b)} Astronomical sensing
    \end{minipage}\hfill
    \begin{minipage}[t]{0.32\textwidth}
        \centering
        \includegraphics[width=\linewidth,height=0.20\textheight,keepaspectratio]{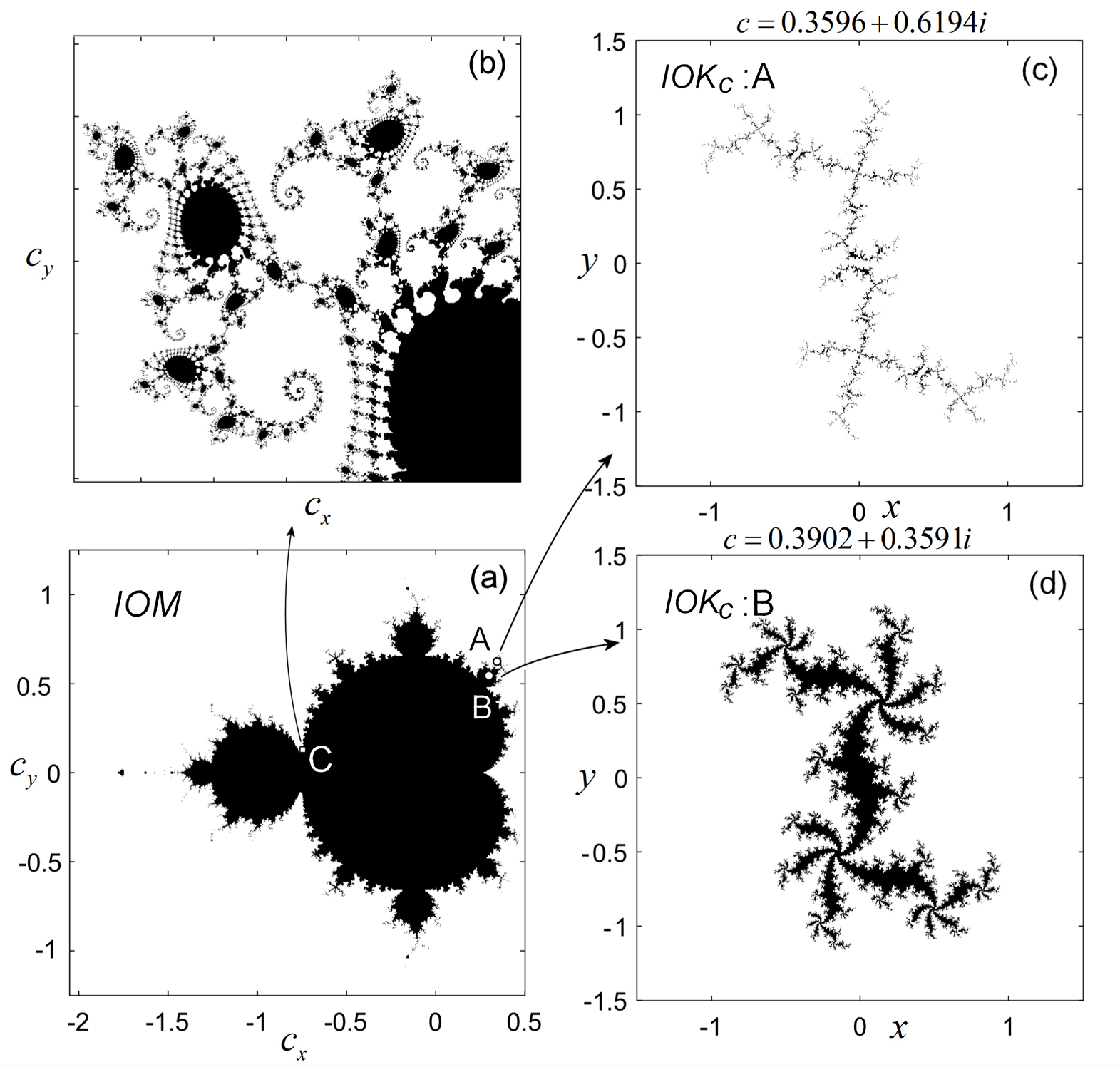}
        \par\smallskip\small\textbf{(c)} Mathematical principle
    \end{minipage}

    \vspace{0.5em}
    \begin{minipage}[t]{0.32\textwidth}
        \centering
        \includegraphics[width=\linewidth,height=0.20\textheight,keepaspectratio]{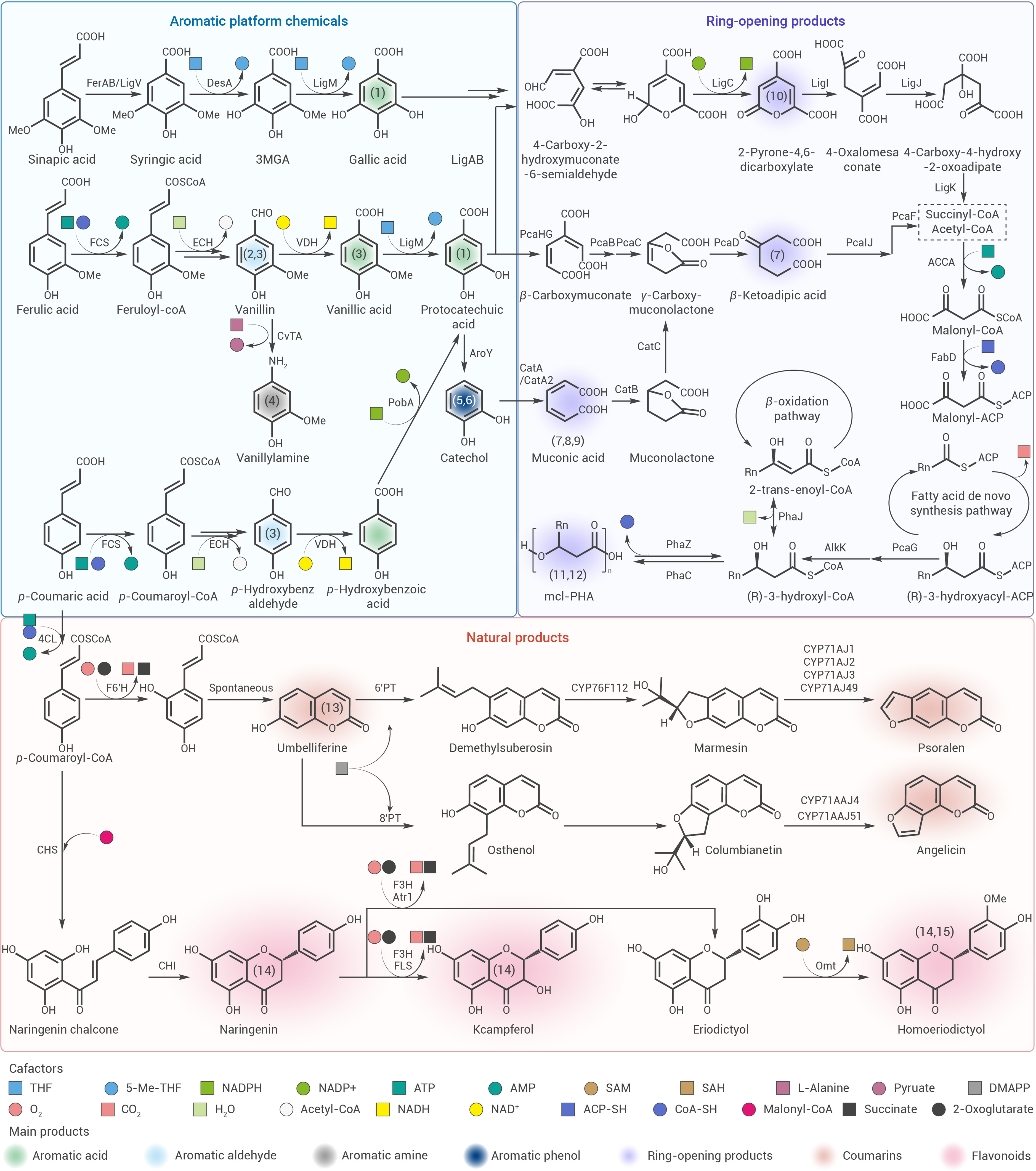}
        \par\smallskip\small\textbf{(d)} Chemical structure and reaction
    \end{minipage}\hfill
    \begin{minipage}[t]{0.32\textwidth}
        \centering
        \includegraphics[width=\linewidth,height=0.20\textheight,keepaspectratio]{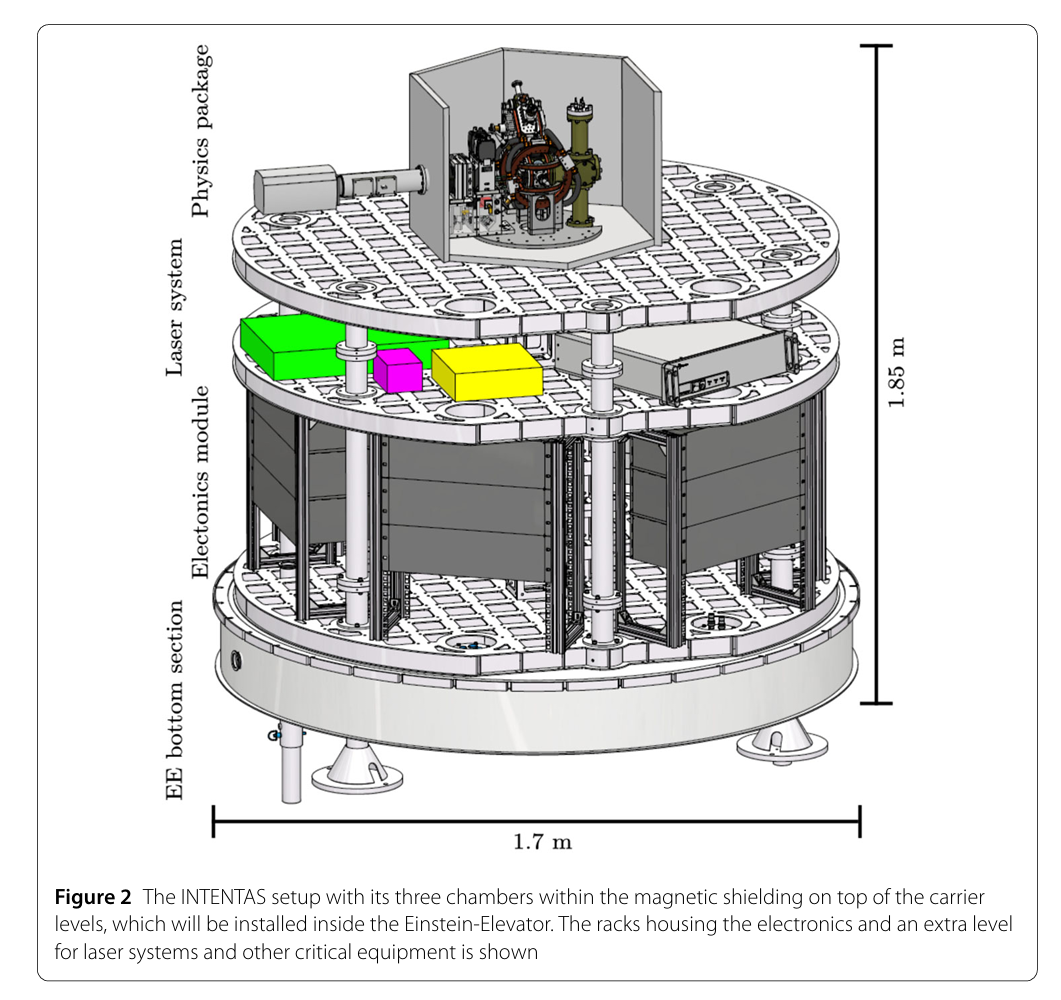}
        \par\smallskip\small\textbf{(e)} Mechanical CAD
    \end{minipage}\hfill
    \begin{minipage}[t]{0.32\textwidth}
        \centering
        \includegraphics[width=\linewidth,height=0.20\textheight,keepaspectratio]{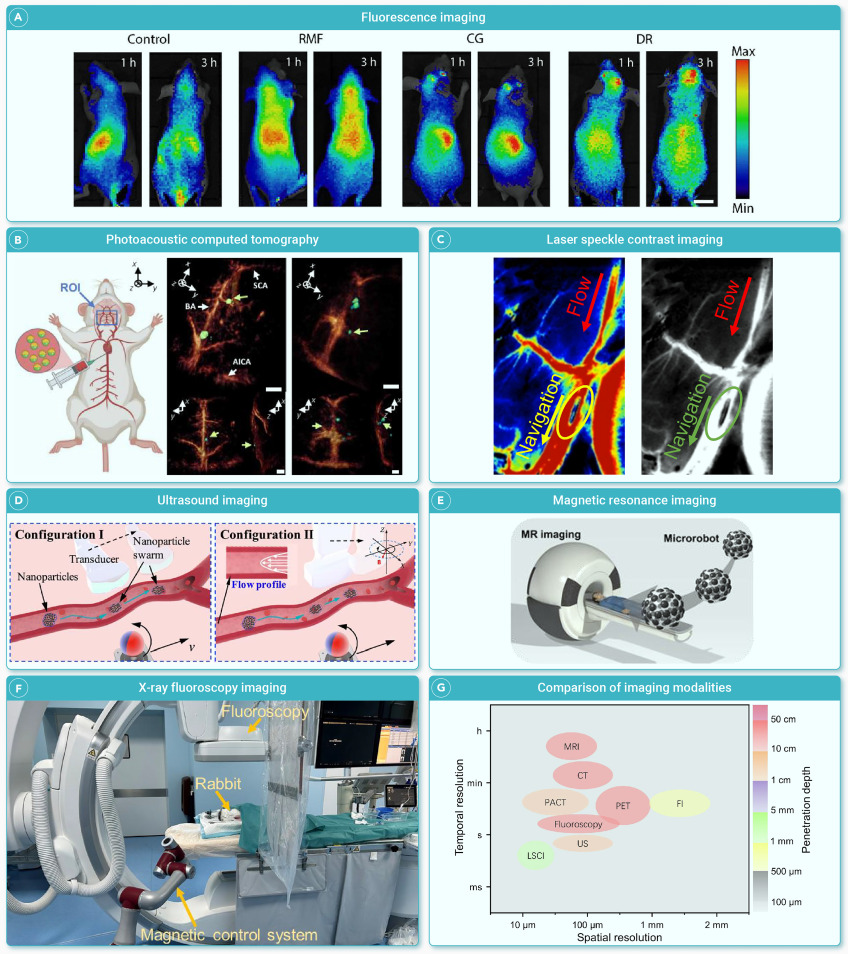}
        \par\smallskip\small\textbf{(f)} Photos and imaging panels
    \end{minipage}
    \caption{Representative examples of categories excluded from end-to-end generation by agents.}
    \label{fig:excluded-category-examples}
\end{figure}

The filtering step in Section~\ref{sec:the_scidiagbench_benchmark} removes diagram categories whose correctness depends on exact numerical, geometric, chemical, spatial, or optical fidelity. Figure~\ref{fig:excluded-category-examples} combines the six representative examples supplied with the dataset materials. They make the filtering decision concrete but are not benchmark instances. The source images are only scaled to fit the composite layout; no panel is cropped, recolored, annotated, or otherwise altered.

\newcommand{\ccbylicense}{\href{https://creativecommons.org/licenses/by/4.0/}{CC BY 4.0}}
\newcommand{\ccbyncndlicense}{\href{https://creativecommons.org/licenses/by-nc-nd/4.0/}{CC BY-NC-ND 4.0}}

\begingroup
\small
\paragraph{Sources and permissions.}
Panel~(a) is reproduced from \citet{chalouan2023cgps}, DOI: \href{https://doi.org/10.3390/s23104846}{10.3390/s23104846}, under the \ccbylicense. Panel~(b) is reproduced unchanged from Figure~1 of \citet{li2025haishao1}, DOI: \href{https://doi.org/10.1016/j.xinn.2025.100949}{10.1016/j.xinn.2025.100949}, under the \ccbyncndlicense. Panel~(c) is reproduced from Figure~1 of \citet{danca2024mandelbrot}, DOI: \href{https://doi.org/10.3390/fractalfract8010069}{10.3390/fractalfract8010069}, under the \ccbylicense. Panel~(d) is reproduced from Figure~2 of \citet{liu2025lignin}, DOI: \href{https://doi.org/10.1016/j.xinn.2025.100993}{10.1016/j.xinn.2025.100993}, under the \ccbylicense. Panel~(e) is reproduced from Figure~2 of \citet{anton2025intentas}, DOI: \href{https://doi.org/10.1140/epjqt/s40507-025-00330-9}{10.1140/epjqt/s40507-025-00330-9}, under the \ccbylicense; the source article does not identify this panel as third-party material. Panel~(f) is reproduced from Figure~4 of \citet{wang2025untethered}, DOI: \href{https://doi.org/10.1016/j.xinn.2025.100874}{10.1016/j.xinn.2025.100874}, under the \ccbyncndlicense. Panels~(b) and~(f) must remain unmodified and are subject to the non-commercial and no-derivatives conditions of their license. Any future modification to a CC BY panel should be identified as an adaptation.

\endgroup

\subsection{Potential Training-Data Contamination Filtering}
\label{sec:appendix-contamination-filtering}

\begin{figure}[H]
    \centering
    \begin{minipage}[t]{0.49\textwidth}
        \centering
        \vspace{0pt}
        \includegraphics[width=\linewidth]{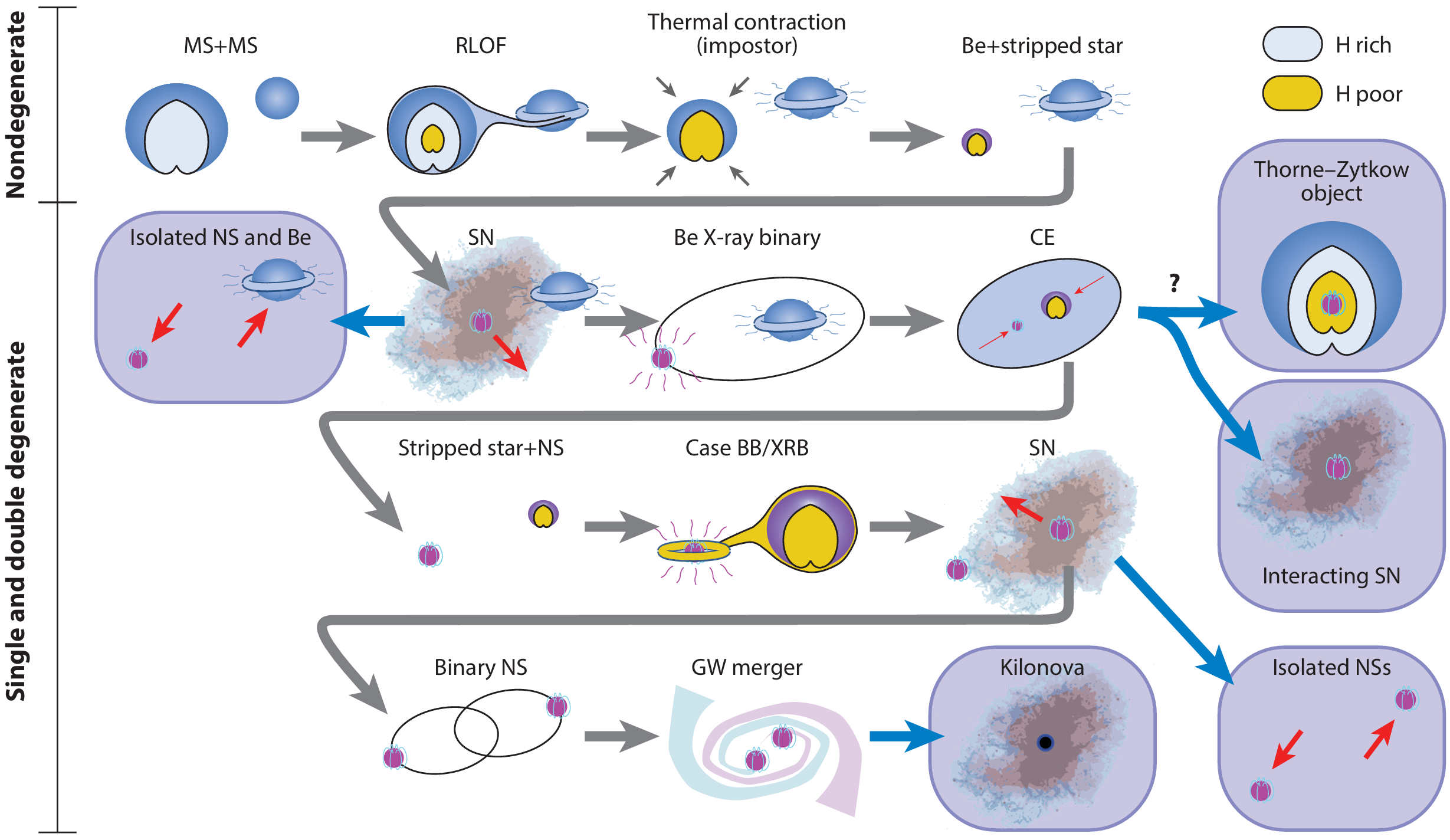}
        \par\smallskip\small\textbf{(a)} Reference diagram
    \end{minipage}\hfill
    \begin{minipage}[t]{0.49\textwidth}
        \centering
        \vspace{0pt}
        \includegraphics[width=\linewidth]{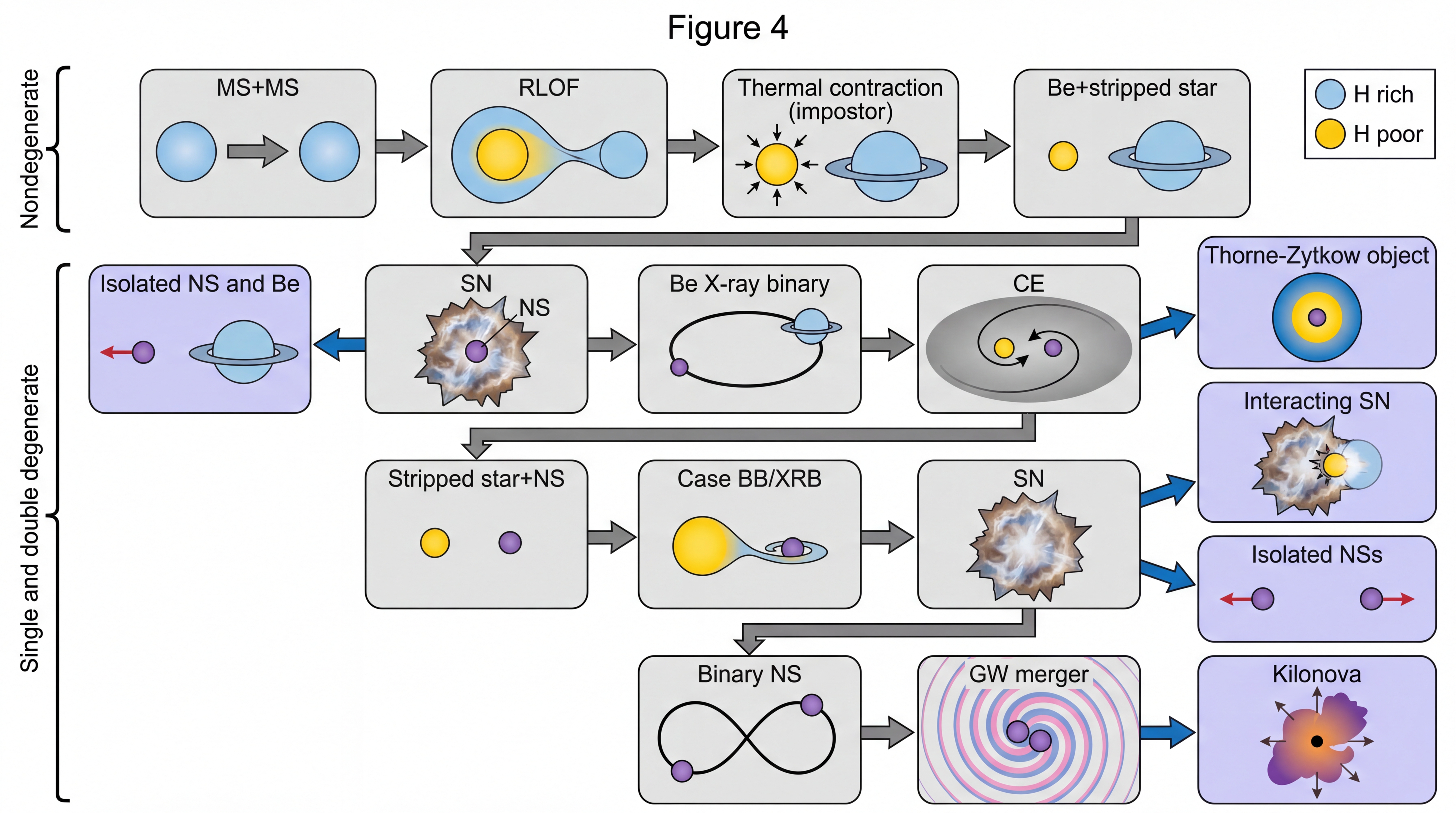}
        \par\smallskip\small\textbf{(b)} Vanilla-generated diagram
    \end{minipage}
    \caption{Example illustrating potential training-data contamination filtering. The reference diagram (a) and Vanilla-generated output (b) share a similar organization of evolutionary stages and branches, motivating precautionary exclusion during manual screening. Both images are shown without cropping or other content modifications.}
    \label{fig:training-data-contamination-example}
\end{figure}

To mitigate potential training-data contamination, we manually inspect the candidate diagrams and exclude those whose preliminary generation output reproduces the reference diagram with anomalously high similarity. Figure~\ref{fig:training-data-contamination-example} illustrates a reference--generation pair used to exemplify this filtering criterion. Despite differences in rendering style, the Vanilla-generated diagram closely follows the reference's overall layout, labeled evolutionary stages, and branching structure. We treat this similarity as a precautionary filtering signal rather than conclusive evidence of training-data contamination.

\subsection{Dataset Composition}
\label{sec:appendix-dataset-composition}

\begin{figure}[H]
    \centering
    \includegraphics[width=1\linewidth]{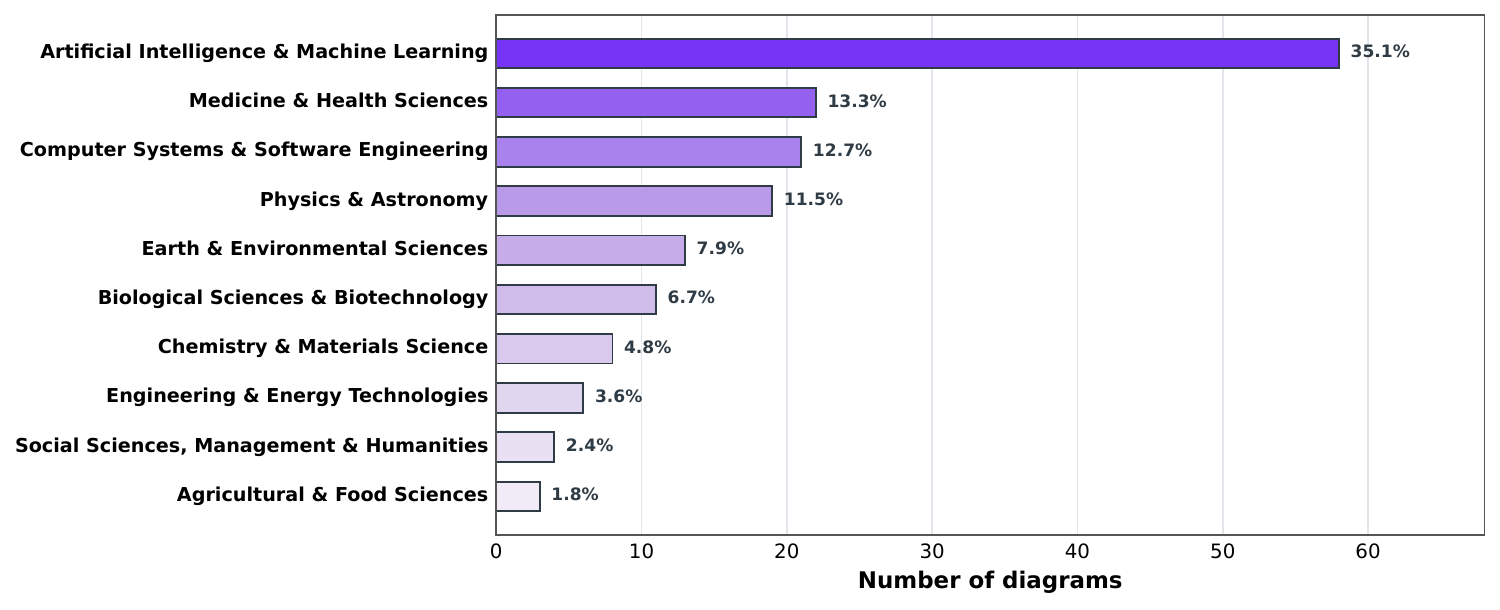}
    \caption{Domain distribution of all 165 diagrams. Bar lengths indicate counts; labels at bar ends indicate percentages.}
    \label{fig:dataset-domain-distribution}
\end{figure}

\section{Detailed Definitions of the 11 Diagram Types}
\label{sec:appendix-diagram-types}

This appendix provides the detailed definition and characteristics of each of the 11 diagram types.

\subsection{Timeline (TL)}
\begin{definitionbox}
A timeline is a visual schematic that arranges information along a chronological axis. It maps the temporal dimension through spatial positioning to illustrate the developmental trajectory, evolutionary trends, or historical progression of a subject.
\end{definitionbox}
\begin{criterionbox}{Characteristics}
\noindent It utilizes a temporal axis as the structural backbone of the diagram, annotated with explicit chronological markers (e.g., specific years such as 2018, 2023, and 2024, or precise months).

\noindent Specific information is conveyed through nodes. These nodes typically incorporate text labels, icons, or logos to signify particular events occurring or milestones achieved at distinct points in time.
\end{criterionbox}

\subsection{Lineage Dendrogram (LD)}
\begin{definitionbox}
A lineage dendrogram is a visual representation illustrating genealogical relationships and evolutionary history. Utilizing a hierarchically distinct branching structure, it depicts how a subject (e.g., biological species, genetic sequences, technological models, or celestial bodies) diverges, proliferates, or iterates from a single common ancestor or root. Driven by the passage of time, environmental selection, or anthropogenic intervention, it delineates the comprehensive historical pathway culminating in the subject's current diversified state.
\end{definitionbox}
\begin{criterionbox}{Characteristics}
\noindent It features a root or origin, denoting the starting point or primordial state of the subject.

\noindent Branches (or edges) represent evolutionary pathways or genetic distances. The length of these lines frequently bears quantitative significance, indicating metrics such as the duration of time or the degree of divergence.
\end{criterionbox}

\subsection{Phased Evolution (PE)}
\begin{definitionbox}
A phased evolution diagram visually represents systemic evolution by segmenting the process into distinct, discrete stages or phases. It primarily focuses on demonstrating qualitative transformations, paradigm shifts, or transitions in environmental states that occur during the subject's evolutionary trajectory. Each phase typically represents a relatively stable state or a specific operational mechanism, with logical progressions or cyclical relationships existing between successive phases.
\end{definitionbox}
\begin{criterionbox}{Characteristics}
\noindent The diagram is partitioned into distinct spatial blocks (e.g., panels, boxes, or tiers), with each block signifying a specific evolutionary stage.

\noindent Within each "stage block," the diagram typically details the micro-structures, chemical/physical reactions, system architectures, or key elements that are unique to that particular phase.
\end{criterionbox}

\subsection{Algorithmic Framework (AF)}
\begin{definitionbox}
An algorithmic framework diagram is a visual schematic that delineates the overall architecture, core components, data flow, and execution logic of an algorithm through a system of graphical symbols. It translates abstract algorithmic models into intuitive, modular blueprints to depict a complete and cohesive framework.
\end{definitionbox}
\begin{criterionbox}{Characteristics}
It demonstrates how an entire system accomplishes complex tasks by orchestrating multiple components. Furthermore, it illustrates the internal structure of the abstract system, encompassing inter-module interactions, engagement with the external environment, and multi-stage process control.
\end{criterionbox}

\subsection{Computer System Architecture (CSA)}
\begin{definitionbox}
A computer system architecture diagram is a visual representation used to abstract and delineate the hardware-software organizational structure, logical component partitioning, and the interaction interfaces and communication mechanisms within a computing system. Utilizing graphical language, it demarcates system boundaries, elucidates the dependencies between different tiers, and illustrates the routing paths of data and control signals across heterogeneous components (e.g., CPUs, GPUs, FPGAs, OS kernels, and user-space applications).
\end{definitionbox}
\begin{criterionbox}{Characteristics}
The system is explicitly partitioned into distinct logical tiers (e.g., hardware layer, driver layer, kernel layer, application layer) or functional modules.
\end{criterionbox}

\subsection{Conceptual Framework (CF)}
\begin{definitionbox}
A conceptual framework diagram is a visual representation that leverages specific visual elements to organize the relationships among the internal components of a research subject in a hierarchical, parallel, or progressive manner. This intuitively illustrates the architecture of a specific research domain or technological trajectory. By mapping the core operational mechanisms, logical workflows, or classification systems through graphical language or theoretical models, it facilitates a rapid comprehension of the "comprehensive overview" or "underlying logic" of the research direction.
\end{definitionbox}
\begin{criterionbox}{Characteristics}
It presents a holistic landscape of a specific research domain or technological architecture, typically characterized by explicit hierarchical or block-based demarcations.
\end{criterionbox}

\subsection{Technical Roadmap (TR)}
\begin{definitionbox}
A technical roadmap is a visual representation of a highly standardized, actionable, and reproducible protocol engineered to achieve specific scientific, engineering, or clinical objectives.
\end{definitionbox}
\begin{criterionbox}{Characteristics}
\noindent Highly Ordered Visual Layout: It demonstrates a strong sense of visual order, utilizing spatial segmentation to articulate temporal and task-based relationships.

\noindent Phased Progression: It typically incorporates explicit phase or stage markers that divide prolonged procedures, thereby facilitating macroscopic progress management.

\noindent Containerization: It employs panels or state-machine concepts to aggregate and categorize parallel processes or attribute-related tasks, effectively mitigating visual clutter.

\noindent Directional Flow: The primary directional flow adheres to conventional "left-to-right" or "top-to-bottom" reading patterns, signifying chronological progression or the primary task trajectory. Inverse flow is strictly restricted to denoting "feedback," "error reporting," or "correction."

\noindent Non-Linear Mechanisms: It may deviate from a singular linear process by incorporating core mechanisms to simulate authentic scientific or engineering decision-making processes, including Branching (path selection predicated on "If-Then" logic, explicitly delineating differentiated execution protocols under varying conditions), Convergence (funnel-based screening predicated on "Filter" logic, illustrating the hierarchical selection process from vast candidate pools to core targets), and Iteration (corrective circuits predicated on "While-Loop" logic, permitting self-correction and feedback via retrograde arrows).

\noindent Multimodal Integration: It may transcend abstract block diagrams by integrating concrete data and states, including State Visualization (clearly illustrating morphological or qualitative transformations occurring within the processed subjects, e.g., data, samples, models, throughout the procedure) and Heterogeneous Data Integration (appending specific reference standards, e.g., anatomical diagrams, code snippets, genetic locus maps, adjacent to the workflow, serving as intuitive benchmarks for the execution of corresponding steps).
\end{criterionbox}

\subsection{Graphical Abstract (GA)}
\begin{definitionbox}
A graphical abstract is a single-panel illustration that provides a structured, intuitive summarization of a scientific paper's core content through highly condensed visual language.
\end{definitionbox}
\begin{criterionbox}{Characteristics}
Serving as the visual counterpart to the textual abstract, it is designed to enable readers to rapidly grasp the research background, underlying mechanisms, principal findings, and conclusions. It functions independently of the main text; readers can comprehend the complete scientific narrative conveyed by the diagram relying solely on its intrinsic elements and annotations, without needing to consult the manuscript for legend explanations.
\end{criterionbox}

\subsection{Experimental Setup Diagram (ESD)}
\begin{definitionbox}
An experimental setup diagram is a visual and abstract representation of a hardware system constructed to validate specific scientific hypotheses or to acquire data.
\end{definitionbox}
\begin{criterionbox}{Characteristics}
\noindent Topology and Flow: It depicts the physical composition and spatial layout of experimental equipment, while explicitly defining the connection topology (physical, optical, or electrical circuits) among various components, as well as the flow logic of energy or information.

\noindent Balanced Abstraction: It oscillates between realism and abstraction by employing primary morphological features or strictly standardized symbols to represent devices, streamlining extraneous details to accentuate structural relationships.
\end{criterionbox}

\subsection{Illustrative Case (IC)}
\begin{definitionbox}
An illustrative case is a visual representation that concretely demonstrates the actual execution process of an algorithm, the detailed reasoning logic of a model, or specific flaws within a system. It achieves this by utilizing concrete, real-world data instances (such as input texts, code snippets, API call parameters, or dialogue contents).
\end{definitionbox}
\begin{criterionbox}{Characteristics}
It employs concrete, real-world data instances (Concrete Examples) as its core medium of representation.
\end{criterionbox}

\subsection{Mechanism Diagram (MD)}
\begin{definitionbox}
A mechanism diagram is a visual representation designed to unveil the specific operational principles underlying a particular phenomenon, function, or event from a microscopic or concrete perspective.
\end{definitionbox}
\begin{criterionbox}{Characteristics}
\noindent It primarily addresses the question "How does this specific event occur? (Principle)" rather than "What kind of system is this? (Structure)", placing a primary emphasis on Process and Function.

\noindent By employing representative elements (e.g., molecules, particles, data streams, force fields), it demonstrates how entities engage in physical, chemical, or informational interactions, explicitly illustrating the outcomes when different elements converge.
\end{criterionbox}

\section{Evaluation Dimensions and Criteria}
\label{sec:appendix-evaluation-dimensions}

Here we provide the detailed definitions and evaluation criteria for the hierarchical evaluation dimensions: Structure, Content, and Aesthetics.

\subsection{Structure}
\begin{definitionbox}
Structure refers to the foundational architectural framework and topological organization of visual elements within a scientific diagram. It constitutes the visual "skeleton" of the diagram, encompassing the spatial layout, logical flow, and hierarchical relationships that govern how information is arranged, grouped, and interconnected.

\end{definitionbox}
\noindent\textbf{Evaluation Criteria}
\begin{criterionbox}{Visual Organizational Layout}
\textbf{Positive Indicators (Strengths):}
\begin{itemize}
    \item Clear Reading Path and Logical Flow: Possesses distinct starting and ending points, ensuring a seamless and intuitive progression of the primary workflow.
    \item Effective Modularization and Grouping: Employs bounding boxes strategically to cluster elements belonging to the same logical tier, thereby reducing cognitive load.
    \item Explicit Hierarchical Relationships: Maintains a clear distinction between the overarching architecture and localized details, rendering parent-child hierarchical dependencies immediately comprehensible.
    \item Clean Topological Routing: Features clear arrow directionality and meticulously avoids unnecessary crossing of connecting lines. When depicting complex feedback loops, it utilizes appropriate curves or detours to preserve overall visual neatness.
    \item Even Spatial Distribution: Achieves a harmonious and well-proportioned distribution of the overall structure across the canvas (via either symmetry or dynamic equilibrium). Maintains a relatively consistent density of information nodes across different regions, complemented by an optimal utilization of white space.
\end{itemize}

\textbf{Negative Indicators (Weaknesses):}
\begin{itemize}
    \item Structural Imbalance and Misplaced Emphasis: Marginalizes core processes to the periphery while centralizing secondary workflows; alternatively, exhibits disproportionate spatial scaling among peer-level modules, thereby misleading readers regarding their relative importance.
    \item Absence of Visual Partitioning: Conflates modules of disparate stages or functions without employing adequate spacing or boundary frames for physical isolation. This results in element overcrowding and potential visual occlusion.
    \item Haphazard Layout and Illogical Arrangement: Exhibits an arbitrary positioning of elements where variations in spatial density lack any underlying structural or semantic rationale. Rather than using intentional clustering or spacing to convey meaning, the layout features random areas of congestion and unjustified empty spaces, resulting in a disorganized visual narrative.
    \item Spatial Detachment and Referential Ambiguity: Positions logically related components—such as text labels, annotations, or legends—at excessive distances from their corresponding graphical objects. This physical separation disrupts visual association, thereby generating referential ambiguity and confusing the reader.
\end{itemize}

\end{criterionbox}
\begin{criterionbox}{Logical Accuracy}
\textbf{Positive Indicators (Strengths):}
\begin{itemize}
    \item Structural Abstraction and In-depth Detail Mining: Effectively distills lengthy and complex text to construct a lucid core visual skeleton while fully extracting and integrating high-value details from the paper (e.g., specific intermediate steps, conditional variables, and micro-mechanisms). Without compromising the readability of the overarching framework, it highly encourages the generation of information-rich, fine-grained diagrams to maximize independent explanatory power.
    \item Precise and Profound Logical Mapping: Ensures the topological structure remains strictly faithful to the original paper's intent. Accurately reconstructs deep-level associations within the text—encompassing not merely basic causal or temporal relationships, but also precisely capturing and visualizing complex feedback loops, conditional branches, and nested hierarchies. This transforms exhaustive textual narratives into a multi-layered, visually articulate logical network.
\end{itemize}

\textbf{Negative Indicators (Weaknesses):}
\begin{itemize}
    \item Detail Overload and Structural Imbalance: Mechanically enumerates redundant nodes or secondary information without adequately distinguishing primary from subordinate elements (i.e., lacking effective information hierarchy management). This induces severe visual clutter and obfuscates the core logical trajectory, thereby negating the diagram's fundamental purpose of elucidating complex information.
    \item Insufficient Granularity and Explanatory Deficit: Exhibits an overly rudimentary structure confined to superficial macroscopic overviews. Omits pivotal transitional steps, core mechanistic branches, or crucial underlying details explicitly articulated in the paper, resulting in vacuous content and incomplete workflows that fail to convey the comprehensive analytical depth of the research.
    \item Structural Hallucination and Logical Misalignment: Presents visual topological relationships that fundamentally conflict with the factual assertions or logical deductions of the paper. Alternatively, it erroneously conflates and associates information nodes that lack any inherent logical intersection during the detail extraction process.
\end{itemize}

\end{criterionbox}
\subsection{Content}
\begin{definitionbox}
Content refers to the concrete, discrete constituent elements utilized to populate the structural framework of a scientific diagram. It represents the specific, localized building blocks—encompassing text labels, scientific nomenclature, alphanumeric data, standard symbols, and individual graphical motifs (such as specific molecular models, cell depictions, or distinct physical components)—that inherently carry factual, domain-specific meaning independent of their spatial arrangement.

\end{definitionbox}
\noindent\textbf{Evaluation Criteria}
\begin{criterionbox}{Visual Element Representation}
\textbf{Positive Indicators (Strengths):}
\begin{itemize}
    \item Precise Semantic Representation: The selected graphical elements, icons, and symbols intuitively and accurately convey their intended scientific meanings through attributes such as color and shape. The element design is highly distilled, entirely devoid of extraneous ornamentation or non-functional details that might impede comprehension.
    \item Adequate Legend Provision: Clear, legible, and appropriately positioned explanatory legends are provided for key abstract elements, abbreviations, non-standard symbols, as well as for color schemes and visual encodings used to differentiate experimental groups or logical classifications. This ensures the self-explanatory capacity of the content.
    \item High-Fidelity Micro-Rendering: Individual visual components (e.g., cellular structures, laboratory apparatuses, molecular models) exhibit sharp edges and accurate proportions, remaining completely free from generative artifacts, blurring, or the 'melting' effects typical of large models. Text and numerical data are rendered in crisp, vectorized, or high-resolution typography, devoid of typographical overlapping or garbled characters.
\end{itemize}

\textbf{Negative Indicators (Weaknesses):}
\begin{itemize}
    \item Redundancy and Semantic Duplication: Elements are subjected to unnecessary repetitive rendering; alternatively, disparate visual entities are employed to represent a singular scientific concept, engendering cognitive ambiguity and misinterpretation.
    \item Generative Morphological Distortion: Localized graphical elements manifest severe morphological anomalies (e.g., erroneous physical linkages in biological structures, logically flawed chemical bond angles) or characteristic AI-generated distortions, thereby undermining the scientific rigor and validity of the content.
    \item Information Deficit and Illegibility: Suboptimal generation quality results in critical text, formula labels, or minute symbols becoming blurred, fragmented, or rendered as unidentifiable garble. Furthermore, the absence of requisite legend guidance for complex systems precludes readers from successfully decoding visual symbols into their underlying scientific concepts.
\end{itemize}

\end{criterionbox}
\begin{criterionbox}{Content Accuracy}
\textbf{Positive Indicators (Strengths):}
\begin{itemize}
    \item Precise and Comprehensive Informational Articulation: Predicated on the exact and unambiguous utilization of technical terminology, data labels, and standardized symbols, the representation thoroughly extracts and delineates intricate details such as specific parameters, experimental conditions, or physical/chemical mechanisms outlined in the paper. The informational elements within the diagram are not solely factually accurate but also manifest a high degree of scientific explanatory capacity.
    \item Terminological Consistency: Abbreviations and variable nomenclatures deployed within the visual elements are strictly consistent with the textual discourse of the paper.
    \item Rigorous Logical Correspondence: The transformation and mapping of informational elements across disparate procedural stages exhibit strict logical coherence, remaining entirely devoid of informational discontinuities or omissions.
    \item High-Value Informational Richness and Depth: The model demonstrates a profound capacity for mining and synthesizing macroscopic information from the paper (encompassing textual narratives, tabular data, and formulaic derivations), effectively translating tacit knowledge or auxiliary data (e.g., specific experimental environments, dosages, temporal intervals) into substantive visual content. Provided that the augmentation of detail does not devolve into excessive clutter that obscures the core scientific value, representations characterized by greater informational density and finer granularity warrant highly positive appraisal.
\end{itemize}

\textbf{Negative Indicators (Weaknesses):}
\begin{itemize}
    \item Insufficiency of Critical Details: There is a failure to effectively extract the requisite parameters, conditions, or mechanistic nuances from the paper that underpin the core logic of the visual representation. As a result, the information remains thin and only lists superficial macroscopic concepts.
    \item Factual, Logical, and Terminological Inaccuracies: The informational elements conveyed contradict the established parameters of the paper or recognized scientific facts. This encompasses profound factual and logical fallacies (e.g., erroneously depicted physical/biological structures, visual representations of data points that contradict the paper's conclusions), as well as surface-level representational inconsistencies and ambiguities. Examples include the utilization of discrepant variable nomenclatures, imprecise or ambiguous technical terminology, erroneous mathematical subscripts or symbols, and the introduction of undefined non-standard abbreviations.
    \item Informational Noise and Irrelevant Elements: Introduced textual descriptions, graphical motifs, or specific data points bear no relevance to the localized thematic context; alternatively, the unrestrained over-accumulation of trivial details engenders a diffusion of visual focus, causing the valuable core content to be submerged beneath extraneous informational noise.
\end{itemize}

\end{criterionbox}
\subsection{Aesthetics}
\begin{definitionbox}
Aesthetics refers to the stylistic articulation, visual presentation, and overall graphic design quality of a scientific diagram. It constitutes the visual 'presentation layer' superimposed upon the structural framework and content elements, encompassing deliberate choices in color palettes (including harmony and contrast), typographic styling (font types and legibility), line weights, gradients, textural details, and holistic visual polish.

\end{definitionbox}
\noindent\textbf{Evaluation Criteria}
\begin{criterionbox}{Stylistic Execution and Visual Polish}
\textbf{Positive Indicators (Strengths):}
\begin{itemize}
    \item Harmonious and Functional Color Palettes: The application of color is both visually unified and functionally deliberate. Colors are appropriately utilized to clearly delineate distinct data modules, guide the viewer's visual flow, or emphasize critical scientific conclusions. The color selection aligns with professional scientific publishing standards and ensures good contrast and accessibility.
    \item Coherent Visual Language: The overall stylistic execution demonstrates a high degree of consistency. The diagram maintains a unified visual language (e.g., strictly adhering to either a 2D flat design or a consistent 3D rendering style) across all independent content elements. Graphical properties—such as stroke weights, corner radii, arrow typologies, and shading techniques—are standardized throughout, preventing any discordant patchwork effect.
    \item Typographic Legibility: Typography is scaled appropriately, and font properties (e.g., font weights, italics) are judiciously applied to establish a clear visual hierarchy and ensure optimal readability without straining the viewer's eyes.
\end{itemize}

\textbf{Negative Indicators (Weaknesses):}
\begin{itemize}
    \item Inharmonious or Chaotic Color Schemes: Exhibits extensive use of highly saturated, fluorescent, or visually clashing color palettes. A distinct lack of color logic is evident (e.g., utilizing dark backgrounds that inadvertently obscure foreground text and reduce contrast).
    \item Fragmented or Collage-like Style: Displays a discordant integration of source materials or AI-generated segments, such as the jarring juxtaposition of flat (2D) and volumetric (3D) graphical styles within the same workflow, resulting in a compromised and fractured visual consistency.
    \item Visual Noise and Unnecessary Decoration: Introduces superficial, non-functional decorative elements (e.g., superfluous drop shadows, exaggerated 3D lighting/glow effects, or overly complex background textures) that obscure primary data and distract readers from the core scientific message, thereby diminishing the academic rigor and seriousness of the diagram.
    \item Disproportionate Scaling and Distortion: Shows inappropriate aspect ratios, wherein critical graphical elements, icons, or typographic components are erroneously stretched, compressed, or otherwise subjected to spatial distortion to fit a layout, sacrificing aesthetic proportion.
\end{itemize}
\end{criterionbox}

\section{Type-Specific Evaluation Weights}
\label{sec:appendix-type-specific-weights}

Table~\ref{tab:type-specific-dimension-weights} reports the top-level dimension weights $W_{d,\tau}$ used in Equation~\ref{eq:weighted-score}. As described in Section~\ref{sec:evaluation-framework}, these weights are generated by Gemini 3.1 Pro and averaged over five independent trials, with the three weights summing to one for each diagram type. Technical Roadmap assigns a higher weight to Structure (0.50) than to Content (0.35), whereas Illustrative Case places greater emphasis on Content (0.50) than on Structure (0.35).

\begin{table}[h]
\centering
\caption{Type-specific top-level dimension weights $W_{d,\tau}$. Each column sums to one. Type abbreviations follow Table~\ref{tab:main-results-types}.}
\label{tab:type-specific-dimension-weights}
\vspace{\baselineskip}
\begingroup
\footnotesize
\setlength{\tabcolsep}{4pt}
\renewcommand{\arraystretch}{1.25}
\begin{tabular}{@{}lccccccccccc@{}}
\toprule
Dimension  & TL   & LD   & PE   & AF   & CSA  & CF   & TR   & GA   & ESD  & IC   & MD   \\
\midrule
Structure  & 0.45 & 0.46 & 0.45 & 0.50 & 0.50 & 0.52 & 0.50 & 0.41 & 0.46 & 0.35 & 0.45 \\
Content    & 0.40 & 0.39 & 0.40 & 0.35 & 0.35 & 0.33 & 0.35 & 0.40 & 0.38 & 0.50 & 0.40 \\
Aesthetics & 0.15 & 0.15 & 0.15 & 0.15 & 0.15 & 0.15 & 0.15 & 0.19 & 0.16 & 0.15 & 0.15 \\
\bottomrule
\end{tabular}
\endgroup
\end{table}

\section{Human Annotation Protocol}
\label{sec:appendix-human-annotations}

\begin{figure}[H]
    \centering
    \includegraphics[width=\textwidth]{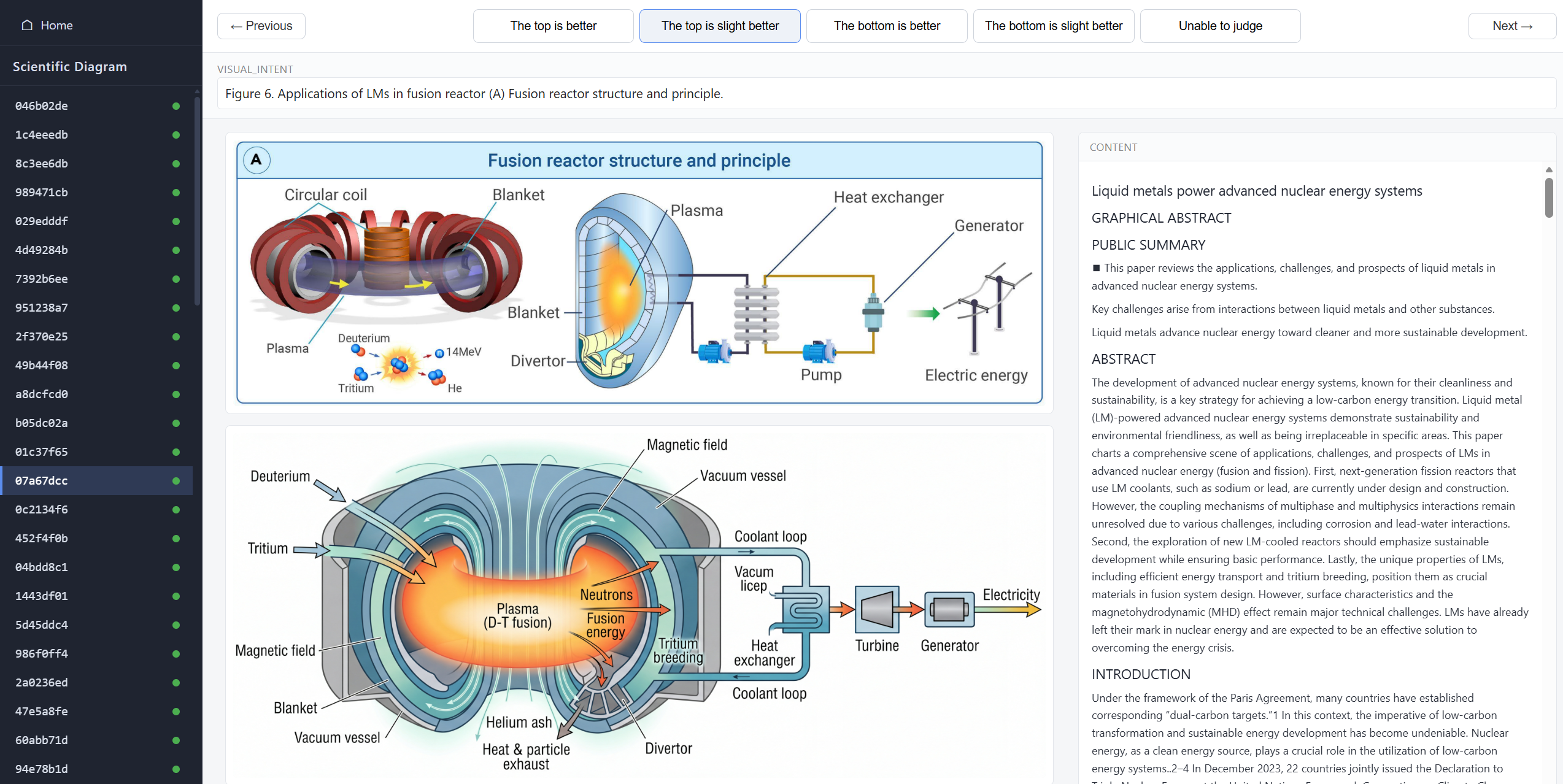}
    \caption{Human annotation interface used to compare a generated diagram with its reference. The screenshot is reproduced unchanged and includes Figure~6A from Lin Zhang, Chang Deng, Xu Ji, and Xiaojing Liu, ``Liquid metals power advanced nuclear energy systems,'' \emph{The Innovation}~6, 100959 (2025), DOI: \href{https://doi.org/10.1016/j.xinn.2025.100959}{10.1016/j.xinn.2025.100959}. The source figure is reproduced under the \href{https://creativecommons.org/licenses/by-nc-nd/4.0/}{CC BY-NC-ND 4.0} license; no changes were made.}
    \label{fig:human-annotation-interface}
\end{figure}

This appendix specifies the protocol used to obtain the human preference labels reported in Section~\ref{sec:exp-setup}. The annotation task compared a generated diagram with its human-created reference for the same paper and target caption. Annotators were shown the pair in a web interface and were not given any automatic evaluation score or method identity. The interface used for this task is shown in Figure~\ref{fig:human-annotation-interface}.

\paragraph{Five-point preference scale.}
Annotators selected one of five ordered responses: ``top is better'', ``top is slightly better'', ``unable to judge'', ``bottom is slightly better'', or ``bottom is better''. The two ``top'' responses correspond to a preference for the generated diagram when it was displayed in the top position; the two ``bottom'' responses correspond to a preference for the reference when it was displayed in the bottom position. The display order was anonymized, and labels were mapped back to generated-versus-reference identities before aggregation. We encode the responses as $+2$, $+1$, $0$, $-1$, and $-2$, respectively, after this identity mapping; the sign denotes direction and the magnitude denotes preference strength.

\paragraph{Independent annotation and strict aggregation.}
Each of the 110 Vanilla--reference pairs was labeled independently by three annotators. For each instance $i$, let $z_{i,r}\in\{-2,-1,0,+1,+2\}$ be annotator $r$'s identity-mapped response. We first assign a directional label $d_{i,r}=\operatorname{sign}(z_{i,r})$: generated-preferred, reference-preferred, or unable to judge. An instance enters the high-agreement set only when at least two annotators provide directionally consistent judgments, and no annotator provides a conflicting directional judgment. Thus, instances lacking at least two consistent directions or containing directional conflicts are excluded. This rule excludes five direction-conflict instances and retains 105 instances, comprising 35 generated-preferred and 70 reference-preferred cases.

For every retained instance, the human preference-strength label is the median of the three signed ordinal responses,
\begin{equation}
    h_i = \operatorname{median}(z_{i,1},z_{i,2},z_{i,3}),
    \qquad i\in\mathcal{H},
\end{equation}
where $\mathcal{H}$ is the 105-instance high-agreement set. The median preserves the ordinal nature of the scale and prevents one annotator's extreme strength judgment from determining the instance label. Directional analyses use $\operatorname{sign}(h_i)$, whereas rank-correlation analyses use $h_i$ itself. No automatic score is used in either filtering or aggregation.

\paragraph{Agreement statistics.}
Fleiss' $\kappa$ is computed on the three-way directional labels over all 110 instances. Ordinal Krippendorff's $\alpha$ is computed from the five-level responses using their ordinal distance. Confidence intervals are obtained by resampling instances, preserving the three annotations within each instance; the resulting estimates are Fleiss' $\kappa=0.730$ (95\% CI $[0.630,0.821]$) and ordinal Krippendorff's $\alpha=0.799$ (95\% CI $[0.744,0.843]$).

\section{Qualitative Case Study}
\label{sec:appendix-qualitative-case}

\begin{figure}[H]
    \centering
    \begin{minipage}[b]{0.31\textwidth}
        \centering
        \includegraphics[width=\linewidth,height=0.15\textheight,keepaspectratio]{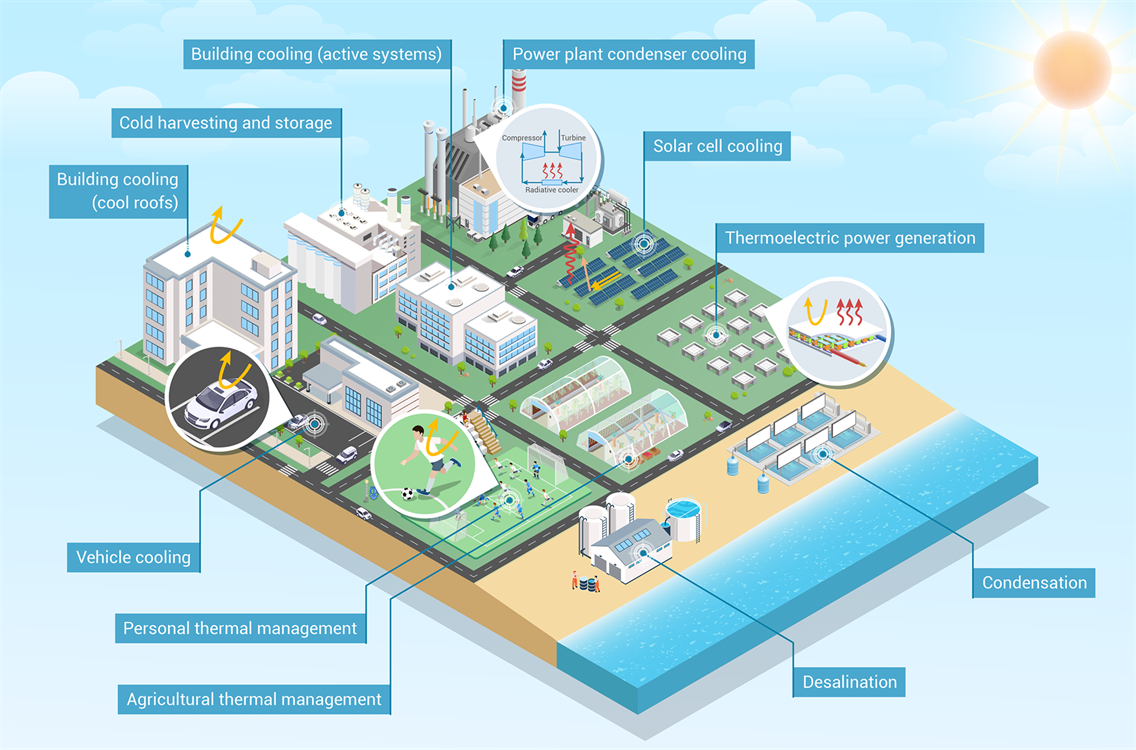}
        \par\smallskip\small\textbf{(a)} Human-created reference
    \end{minipage}\hspace{0.015\textwidth}
    \begin{minipage}[b]{0.31\textwidth}
        \centering
        \includegraphics[width=\linewidth,height=0.15\textheight,keepaspectratio]{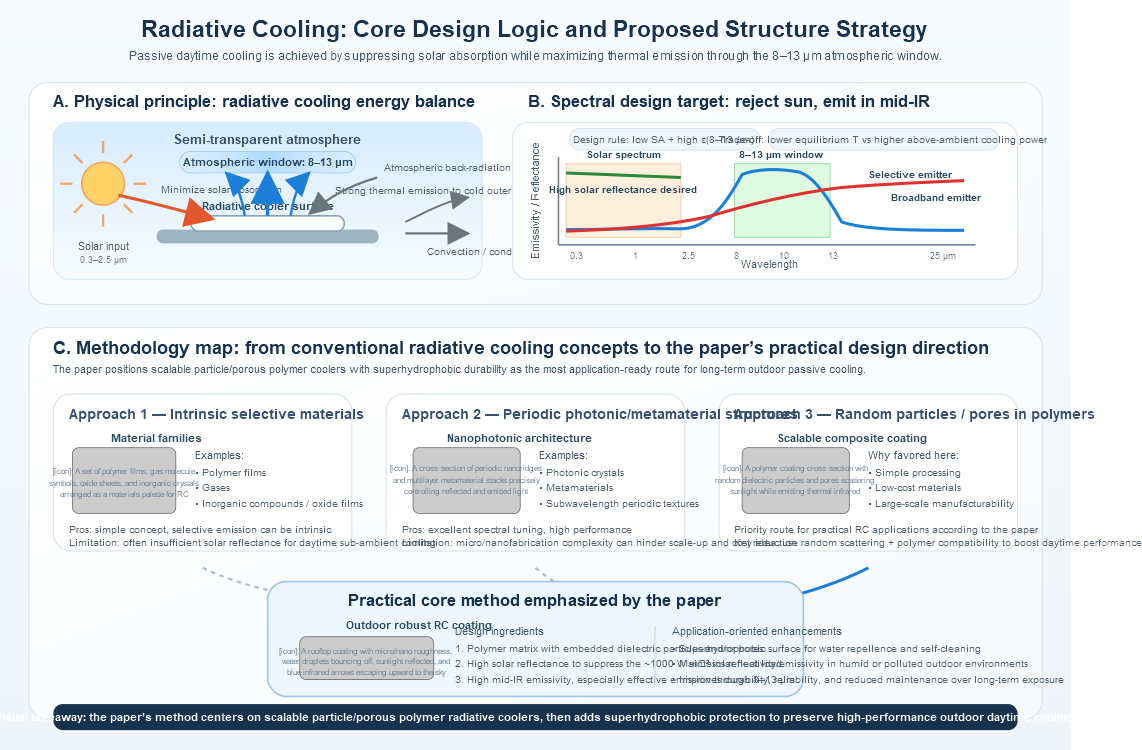}
        \par\smallskip\small\textbf{(b)} AutoFigure output
    \end{minipage}\hspace{0.015\textwidth}
    \begin{minipage}[b]{0.31\textwidth}
        \centering
        \includegraphics[width=\linewidth,height=0.15\textheight,keepaspectratio]{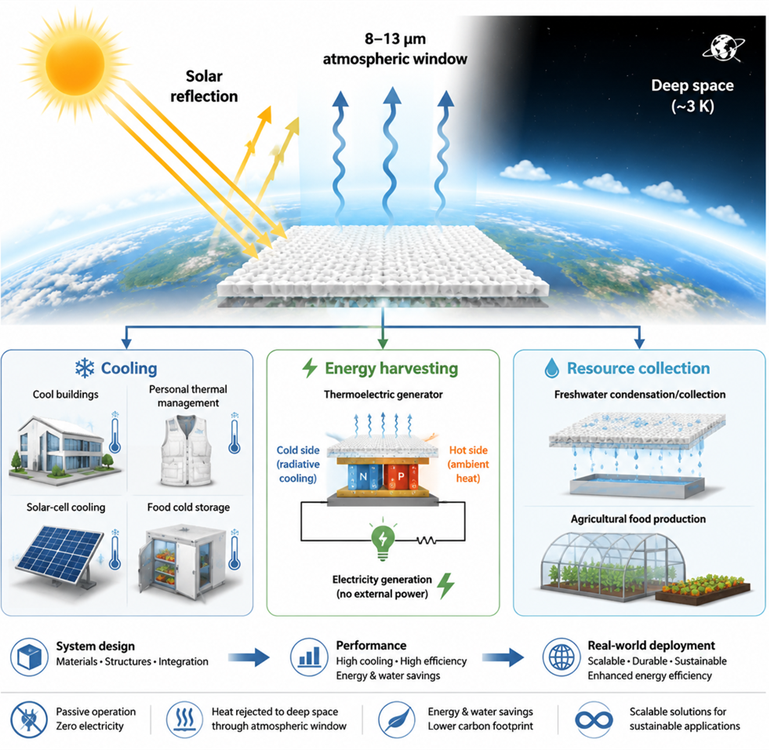}
        \par\smallskip\small\textbf{(c)} Paper2Figure output
    \end{minipage}

    \par\vspace{0.8em}
    \begin{minipage}[b]{0.31\textwidth}
        \centering
        \includegraphics[width=\linewidth,height=0.15\textheight,keepaspectratio]{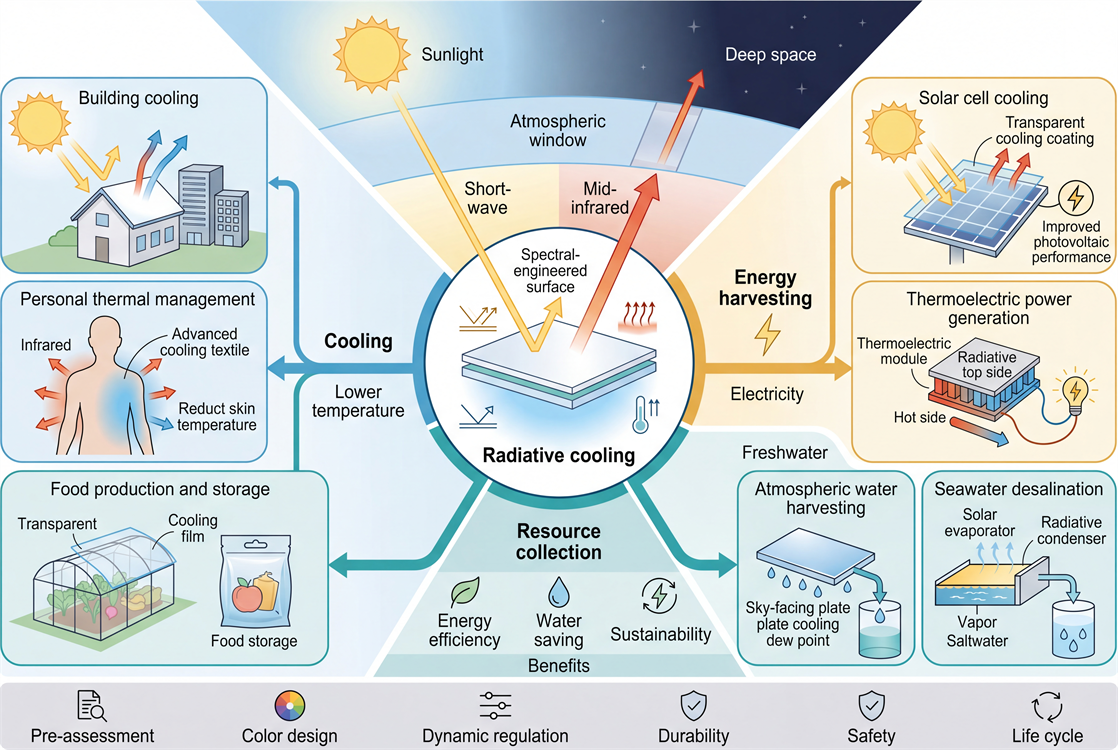}
        \par\smallskip\small\textbf{(d)} Vanilla output
    \end{minipage}\hspace{0.03\textwidth}%
    \begin{minipage}[b]{0.31\textwidth}
        \centering
        \includegraphics[width=\linewidth,height=0.15\textheight,keepaspectratio]{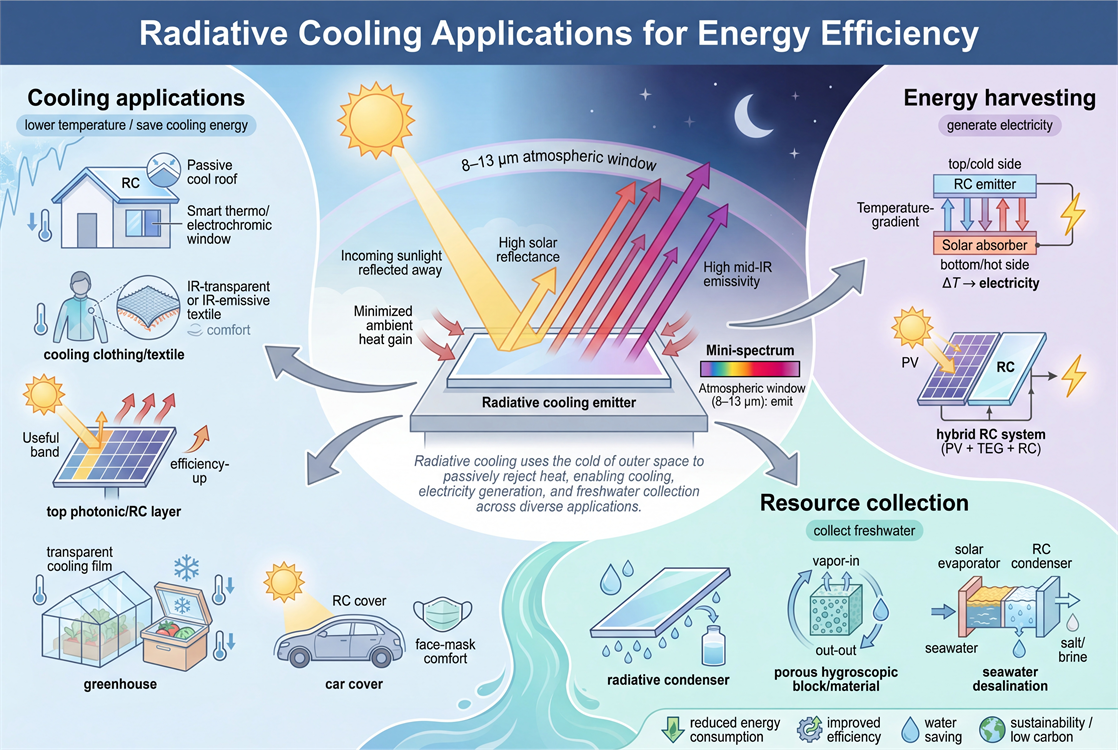}
        \par\smallskip\small\textbf{(e)} PaperBanana output
    \end{minipage}
    \caption{Top row: (a) the original human-created reference, (b) AutoFigure output, and (c) Paper2Figure output. Bottom row: (d) Vanilla output and (e) PaperBanana output.}
    \label{fig:qualitative-case}
\end{figure}

Figure~\ref{fig:qualitative-case} presents one example instance from the Illustrative Case category.

\section{Per-Category Dimension-Level Results}
\label{sec:appendix-category-dimensions}

\begin{table}[H]
\caption{Per-category dimension-level results for PaperBanana and Vanilla. Structure and Content rows are the weighted top-level scores; VOL, LA, VER, and CA are their corresponding sub-dimensions.}
\label{tab:appendix-category-dimensions}
\centering
\vspace{\baselineskip}
\scriptsize
\renewcommand{\arraystretch}{0.95}
\resizebox{\textwidth}{!}{%
\begin{tabular}{llcccccccccccc}
\toprule
\multirow{2}{*}{Method} & \multirow{2}{*}{Dimension} & \multicolumn{11}{c}{Diagram type} & \multirow{2}{*}{Overall} \\
\cmidrule(lr){3-13}
 & & TL & LD & PE & AF & CSA & CF & TR & GA & ESD & IC & MD & \\
\midrule
\multirow{8}{*}{PaperBanana}
 & Structure (weighted) & 0.5379 & 0.4843 & 0.5797 & 0.5632 & 0.5091 & 0.5834 & 0.5243 & 0.6204 & 0.4572 & 0.5037 & 0.4953 & 0.5326 \\
 & \quad VOL & 0.5302 & 0.5086 & 0.5738 & 0.5697 & 0.5076 & 0.5786 & 0.5295 & 0.6186 & 0.4833 & 0.5361 & 0.5176 & 0.5412 \\
 & \quad LA & 0.5459 & 0.4659 & 0.5893 & 0.5553 & 0.5105 & 0.5896 & 0.5174 & 0.6228 & 0.4355 & 0.4764 & 0.4766 & 0.5259 \\
 & Content (weighted) & 0.5002 & 0.4827 & 0.6195 & 0.5574 & 0.4810 & 0.5557 & 0.5432 & 0.6090 & 0.4478 & 0.4586 & 0.4914 & 0.5224 \\
 & \quad VER & 0.5313 & 0.5168 & 0.6306 & 0.5683 & 0.4821 & 0.5371 & 0.5577 & 0.5858 & 0.4685 & 0.4779 & 0.4935 & 0.5318 \\
 & \quad CA & 0.4803 & 0.4529 & 0.6146 & 0.5482 & 0.4798 & 0.5686 & 0.5338 & 0.6311 & 0.4323 & 0.4435 & 0.4885 & 0.5158 \\
 & Aesthetics & 0.5762 & 0.5366 & 0.6020 & 0.6003 & 0.4674 & 0.5564 & 0.5597 & 0.6470 & 0.5058 & 0.5192 & 0.5281 & 0.5544 \\
 & Overall & 0.5272 & 0.4917 & 0.5975 & 0.5665 & 0.4926 & 0.5703 & 0.5361 & 0.6195 & 0.4610 & 0.4824 & 0.4985 & 0.5312 \\
\midrule
\multirow{8}{*}{Vanilla}
 & Structure (weighted) & 0.4515 & 0.3957 & 0.5464 & 0.4531 & 0.4980 & 0.5912 & 0.5003 & 0.6198 & 0.5071 & 0.3675 & 0.4694 & 0.4909 \\
 & \quad VOL & 0.4602 & 0.4171 & 0.5382 & 0.4684 & 0.5022 & 0.5979 & 0.5358 & 0.6193 & 0.5428 & 0.4122 & 0.4973 & 0.5083 \\
 & \quad LA & 0.4418 & 0.3794 & 0.5574 & 0.4362 & 0.4942 & 0.5848 & 0.4654 & 0.6188 & 0.4758 & 0.3301 & 0.4475 & 0.4756 \\
 & Content (weighted) & 0.3970 & 0.4158 & 0.5772 & 0.4565 & 0.4860 & 0.5687 & 0.4976 & 0.6087 & 0.5133 & 0.3566 & 0.4351 & 0.4829 \\
 & \quad VER & 0.4433 & 0.4683 & 0.5851 & 0.4859 & 0.4733 & 0.5448 & 0.5494 & 0.5876 & 0.5465 & 0.4235 & 0.4548 & 0.5057 \\
 & \quad CA & 0.3631 & 0.3648 & 0.5736 & 0.4359 & 0.4947 & 0.5851 & 0.4538 & 0.6282 & 0.4910 & 0.3050 & 0.4202 & 0.4650 \\
 & Aesthetics & 0.5123 & 0.4652 & 0.5847 & 0.5368 & 0.5040 & 0.5833 & 0.5638 & 0.6547 & 0.6068 & 0.4484 & 0.4725 & 0.5393 \\
 & Overall & 0.4406 & 0.4168 & 0.5626 & 0.4670 & 0.4947 & 0.5820 & 0.5094 & 0.6228 & 0.5258 & 0.3740 & 0.4566 & 0.4957 \\
\bottomrule
\end{tabular}}
\end{table}

Table~\ref{tab:appendix-category-dimensions} reports the complete per-category breakdown for PaperBanana and Vanilla. In addition to the weighted Structure and Content scores, we report their Visual Organizational Layout (VOL), Logical Accuracy (LA), Visual Element Representation (VER), and Content Accuracy (CA) sub-dimensions, together with Aesthetics and the overall score. All entries are relative scores against the human-created reference, so 0.5 denotes human parity. Category abbreviations follow Table~\ref{tab:main-results-types}.

\paragraph{Analysis.}
The breakdown clarifies why the aggregate advantage of the more elaborate agent is not uniform. PaperBanana outperforms Vanilla on seven of the eleven categories, but Vanilla remains better on Computer System Architecture (CSA), Conceptual Framework (CF), Graphical Abstract (GA), and Experimental Setup Diagram (ESD). The ESD gap is especially pronounced: PaperBanana's overall score is 0.4610 versus 0.5258 for Vanilla, and it trails Vanilla in every reported dimension, including Structure (0.4572 vs.~0.5071), Content (0.4478 vs.~0.5133), and Aesthetics (0.5058 vs.~0.6068). In particular, Content Accuracy is 0.4323 for PaperBanana compared with 0.4910 for Vanilla. CSA shows a milder version of the same failure mode: PaperBanana has slightly stronger Structure (0.5091 vs.~0.4980), but lower Content (0.4810 vs.~0.4860), Content Accuracy (0.4798 vs.~0.4947), and Aesthetics (0.4674 vs.~0.5040), leaving the overall scores nearly tied (0.4926 vs.~0.4947). The fact that Content Accuracy is lower for PaperBanana on both ESD and CSA is consistent with the possibility that explicit intermediate content planning can discard details or impose an overly rigid structure on categories that benefit from holistic generation. This table-level evidence is diagnostic rather than causal, and does not by itself identify which planning component is responsible.

The difficult categories also reveal a ceiling shared by current systems. PaperBanana improves over Vanilla on Lineage Dendrogram (0.4917 vs.~0.4168), Illustrative Case (0.4824 vs.~0.3740), and Mechanism Diagram (0.4985 vs.~0.4566), yet remains below the 0.5 human-parity reference in all three. Their accuracy sub-dimensions are particularly limiting: PaperBanana reaches only 0.4529/0.4659 (CA/LA) on Lineage Dendrogram, 0.4435/0.4764 on Illustrative Case, and 0.4885/0.4766 on Mechanism Diagram. Thus, planning improves relative performance over the minimal baseline, but does not reliably solve precise topology, concrete entity depiction, or causal-relation fidelity. Across these categories, the recurring gap between visual organization or element representation and the corresponding accuracy score suggests that diagrams can look plausible while still communicating incomplete or incorrect scientific relations.

\section{Underlying-Model Sensitivity}
\label{sec:underlying-model-sensitivity}

We assess backbone sensitivity within PaperBanana by replacing either its VLM or its Multimodal Generative Model while holding the other component fixed. The VLM variants are Gemini 3.1 Pro~\citep{google2026gemini31pro}, Qwen3.5-Plus~\citep{alibaba2026qwen35plus}, and Seed 2.0 Pro~\citep{bytedance2026seed2}; the multimodal generative model variants are GPT-Image-2~\citep{openai2026gptimage2}, Qwen-Image-2.0-Pro~\citep{alibaba2026qwenimage20pro}, and Seedream 5~\citep{bytedance2026seedream5}.

\begin{table}[H]
\caption{PaperBanana results under different underlying models. The default configuration pairs GPT-5.4 with Gemini 3 Pro Image; each variant replaces one component while holding the other fixed. Type abbreviations follow Table~\ref{tab:main-results-types}.}
\label{tab:underlying-model-results}
\centering
\vspace{\baselineskip}
\resizebox{\textwidth}{!}{%
\begin{tabular}{lcccccccccccc}
\toprule
Model & TL & LD & PE & AF & CSA & CF & TR & GA & ESD & IC & MD & Overall \\
\midrule
Default & 0.5272 & 0.4917 & \textbf{0.5975} & \textbf{0.5665} & 0.4926 & 0.5703 & \textbf{0.5361} & \textbf{0.6195} & 0.4610 & \textbf{0.4824} & \textbf{0.4985} & \textbf{0.5312} \\
\midrule
\multicolumn{13}{l}{\textit{VLM variants}} \\
Gemini 3.1 Pro & 0.4439 & 0.3824 & 0.5617 & 0.5380 & \textbf{0.4992} & \textbf{0.5835} & 0.4835 & 0.5699 & \textbf{0.5200} & 0.4751 & 0.4204 & 0.4980 \\
Qwen3.5-Plus & 0.4723 & 0.4103 & 0.5582 & 0.4905 & 0.4895 & 0.5721 & 0.4773 & 0.5850 & 0.4465 & 0.4750 & 0.4681 & 0.4950 \\
Seed 2.0 Pro & \textbf{0.5277} & \textbf{0.5108} & 0.5758 & 0.4620 & 0.4848 & 0.5614 & 0.4597 & 0.5808 & 0.4371 & 0.4455 & 0.4624 & 0.5007 \\
\midrule
\multicolumn{13}{l}{\textit{Multimodal Generative Model variants}} \\
GPT-Image-2 & 0.4627 & 0.4914 & 0.5932 & 0.5153 & 0.4845 & 0.5524 & 0.5082 & 0.6140 & 0.4493 & 0.4787 & 0.4318 & 0.5074 \\
Qwen-Image-2.0-Pro & 0.2626 & 0.2334 & 0.4801 & 0.2739 & 0.3443 & 0.3525 & 0.3546 & 0.3939 & 0.3378 & 0.3071 & 0.2729 & 0.3285 \\
Seedream 5 & 0.2835 & 0.2773 & 0.4055 & 0.2563 & 0.2808 & 0.3144 & 0.2854 & 0.2874 & 0.2458 & 0.2411 & 0.2290 & 0.2824 \\
\bottomrule
\end{tabular}}
\end{table}

\paragraph{VLM Comparison.}
With Gemini 3 Pro Image fixed, overall scores span a relatively narrow interval from 0.4950 to 0.5312. GPT-5.4 attains the highest overall score and leads on six of the 11 diagram types, while Gemini 3.1 Pro and Seed 2.0 Pro achieve the best performance on the remaining five types.

\paragraph{Multimodal Generative Model Comparison.}
With GPT-5.4 fixed, Gemini 3 Pro Image obtains the highest score on every diagram type and overall. GPT-Image-2 is comparatively close in aggregate (0.5074 vs.\ 0.5312), whereas Qwen-Image-2.0-Pro and Seedream 5 yield substantially lower overall scores of 0.3285 and 0.2824.

\section{Ablation Results for GPT-5.4 Judge}
\label{sec:appendix-ablation}

Table~\ref{tab:appendix-judge-framework-validity} presents the ablation results for the evaluation framework using GPT-5.4 as the judge VLM (the results for Gemini 3.1 Pro are provided in Section~\ref{sec:ablation-study}). While GPT-5.4 exhibits similar positive trends for the directional agreement rate (DAR) point estimates when the proposed mechanisms are applied, its overall alignment remains low due to its inherent calibration defects discussed in Section~\ref{sec:evaluation-reliability}.

\begin{table}[H]
\caption{Human-alignment results for the $2\times2$ evaluation-framework analysis on the 105 high-agreement instances using GPT-5.4. Gen. and Ref. recall correspond to human preferences for the generated and reference diagrams, respectively. DAR is the primary endpoint.}
\label{tab:appendix-judge-framework-validity}
\centering
\vspace{\baselineskip}
\footnotesize
\begin{tabular*}{0.9\textwidth}{@{\extracolsep{\fill}}ccccccc@{}}
\toprule
\multicolumn{2}{c}{Experimental setup} & \multicolumn{3}{c}{Directional Agreement Rate} & \multicolumn{2}{c}{Strength Alignment} \\
\cmidrule(lr){1-2} \cmidrule(lr){3-5} \cmidrule(lr){6-7}
TSC & HPA & Overall & Gen.-pref. & Ref.-pref. & Kendall's $\tau_b$ & Spearman's $\rho$ \\
\midrule
-- & -- & 0.4952 & 0.7143 & 0.3857 & 0.0989 & 0.1292 \\
-- & \checkmark & 0.5143 & \textbf{0.8286} & 0.3571 & \textbf{0.2680} & \textbf{0.3444} \\
\checkmark & -- & 0.5143 & 0.7143 & 0.4143 & 0.0625 & 0.0710 \\
\checkmark & \checkmark & \textbf{0.5714} & 0.7143 & \textbf{0.5000} & 0.2277 & 0.2934 \\
\bottomrule
\end{tabular*}
\end{table}

\section{Type-Level Stability Analysis}
\label{sec:appendix-type-level-stability}

This appendix complements the aggregate stability results in Section~\ref{sec:repeated-scoring-stability} with a breakdown across diagram types. We use the same three independent scoring runs on \benchname{}-Test and the same four configurations: ours, w.o. HPA, w.o. TSC, and w.o. HPA\&TSC.

\begin{figure}[H]
    \centering
    \includegraphics[width=\textwidth]{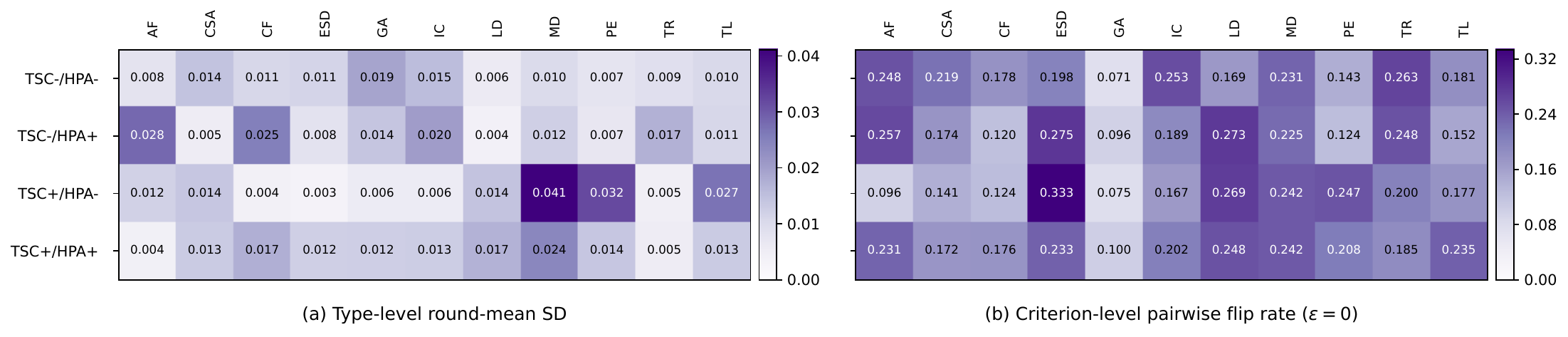}
    \caption{Stability across diagram types. The left panel reports the type-level standard deviation (SD) across three runs, while the right panel reports criterion-level pairwise flip rates with $\varepsilon=0$.}
    \label{fig:category-stability-heatmap}
\end{figure}

Figure~\ref{fig:category-stability-heatmap} reports the standard deviation of type-level mean scores across the three runs and the criterion-level pairwise flip rate for each of the 11 diagram types. As shown in Figure~\ref{fig:category-stability-heatmap}, configurations using only one of TSC or HPA exhibit several pronounced type-specific fluctuations. For example, the TSC-only configuration shows relatively large score variations on Mechanism Diagram, Phased Evolution, and Timeline, whereas the HPA-only configuration exhibits a larger fluctuation on Algorithmic Framework. Experimental Setup Diagram also shows relatively high criterion-level flip rates under several configurations. In comparison, integrating TSC and HPA reduces some of these extreme type-specific variations and yields a more balanced stability profile across diagram types. These observations suggest that, although the complete framework does not uniformly improve every aggregate stability metric, combining TSC and HPA can help mitigate pronounced instability on particular diagram types. This may reflect the complementary roles of type-specific criteria and preference-based alignment in constraining the variability of the judge model across heterogeneous scientific diagrams.

\section{Limitations}
\label{sec:appendix-limitations}

\paragraph{Limited dataset size.}
The benchmark contains 165 high-quality instances. Although these instances cover 11 diagram types and 10 scientific domains, reflecting the diversity of scientific diagrams encountered in practice, the dataset size is currently insufficient to support large-scale training or fine-tuning of models. A natural direction for future work is to use the current high-quality instances as seeds for automated expansion, followed by systematic filtering and human quality control. Such a process could increase coverage while preserving the structural and semantic standards of the benchmark.

\paragraph{Limited coverage of generation improvement.}
This work primarily studies how to evaluate scientific diagram generation. We include a Vanilla agent as a baseline to provide a representative generation system for the evaluation experiments, but do not present a comprehensive method for improving the generation process itself. Designing agents that can better plan, render, and verify diagrams remains an important direction for future research.

\paragraph{Residual reference-preference bias in VLM evaluation.}
Table~\ref{tab:judge-framework-validity} shows that the largest source of VLM evaluation error is the misclassification of reference-preferred instances. In particular, among instances that human annotators judge to be better represented by the reference image, the VLM sometimes assigns preference to the generated image: even with both TSC and HPA, the DAR is $0.8143$ for the Ref.-pref. stratum, corresponding to an error rate of $18.57\%$, compared with a DAR of $0.9429$ and an error rate of $5.71\%$ for the Gen.-pref. stratum. Our evaluation framework suppresses this bias, but does not eliminate it completely.

\end{document}